\documentclass{article}

     \usepackage[final]{neurips_2019}

\usepackage[utf8]{inputenc} % allow utf-8 input
\usepackage[T1]{fontenc}    % use 8-bit T1 fonts
\usepackage{hyperref}       % hyperlinks
\usepackage{url}            % simple URL typesetting
\usepackage{booktabs}       % professional-quality tables
\usepackage{amsfonts}       % blackboard math symbols
\usepackage{nicefrac}       % compact symbols for 1/2, etc.
\usepackage{microtype}      % microtypography

\usepackage{xcolor}
\hypersetup{
    colorlinks,
    linkcolor={black!50!black},
    citecolor={black!50!black},
    urlcolor={blue!20!black}
}

\usepackage[square,numbers, sort&compress]{natbib}
\title{Knowledge Informed Machine Learning using a Weibull-based Loss Function}

\author{
  Tim~von Hahn and Chris K Mechefske\\
  Department of Mechanical and Materials Engineering\\
  Queen's University\\
}
\begin{document}

\maketitle

\begin{abstract}
Machine learning can be enhanced through the integration of external knowledge. This method, called knowledge informed machine learning, is also applicable within the field of Prognostics and Health Management (PHM). In this paper, the various methods of knowledge informed machine learning, from a PHM context, are reviewed with the goal of helping the reader understand the domain. In addition, a knowledge informed machine learning technique is demonstrated, using the common IMS and PRONOSTIA bearing data sets, for remaining useful life (RUL) prediction. Specifically, knowledge is garnered from the field of reliability engineering which is represented through the Weibull distribution. The knowledge is then integrated into a neural network through a novel Weibull-based loss function. A thorough statistical analysis of the Weibull-based loss function is conducted, demonstrating the effectiveness of the method on the PRONOSTIA data set. However, the Weibull-based loss function is less effective on the IMS data set. The results, shortcomings, and benefits of the approach are discussed in length. Finally, all the code is publicly available for the benefit of other researchers.\footnote[1]{\href{https://github.com/tvhahn/weibull-knowledge-informed-ml}{https://github.com/tvhahn/weibull-knowledge-informed-ml}}
\end{abstract}

% \keywords{Tool Wear Monitoring; Deep Learning; Machine Learning; Machinery Health Monitoring; Anomaly Detection; Self-Supervised Learning; Variational Autoencoder}

%######% Introduction %######% 
\section{Introduction}

Machine learning (ML) is a useful tool in developing Prognostics and Health Management (PHM) solutions. However, many modern machine learning techniques center around the creation of data sets, the design of neural networks, and exploiting computational power. Yet, there is much domain knowledge, embedded in well proven equations, models, or methods, that can also be harnessed to improve machine learning applications.

Knowledge informed machine learning is the combination of machine learning techniques with explicit prior knowledge. The prior knowledge exists outside of a traditional machine learning pipeline. As such, feature engineering and signal processing techniques are excluded from the definition of knowledge informed ML. In the context of PHM, knowledge informed ML can improve predictive performance and help overcome obstacles such as limited or poor-quality data \cite{muralidhar2018incorporating}.

\begin{figure}
  \centering
  \includegraphics[width=1.0\linewidth]{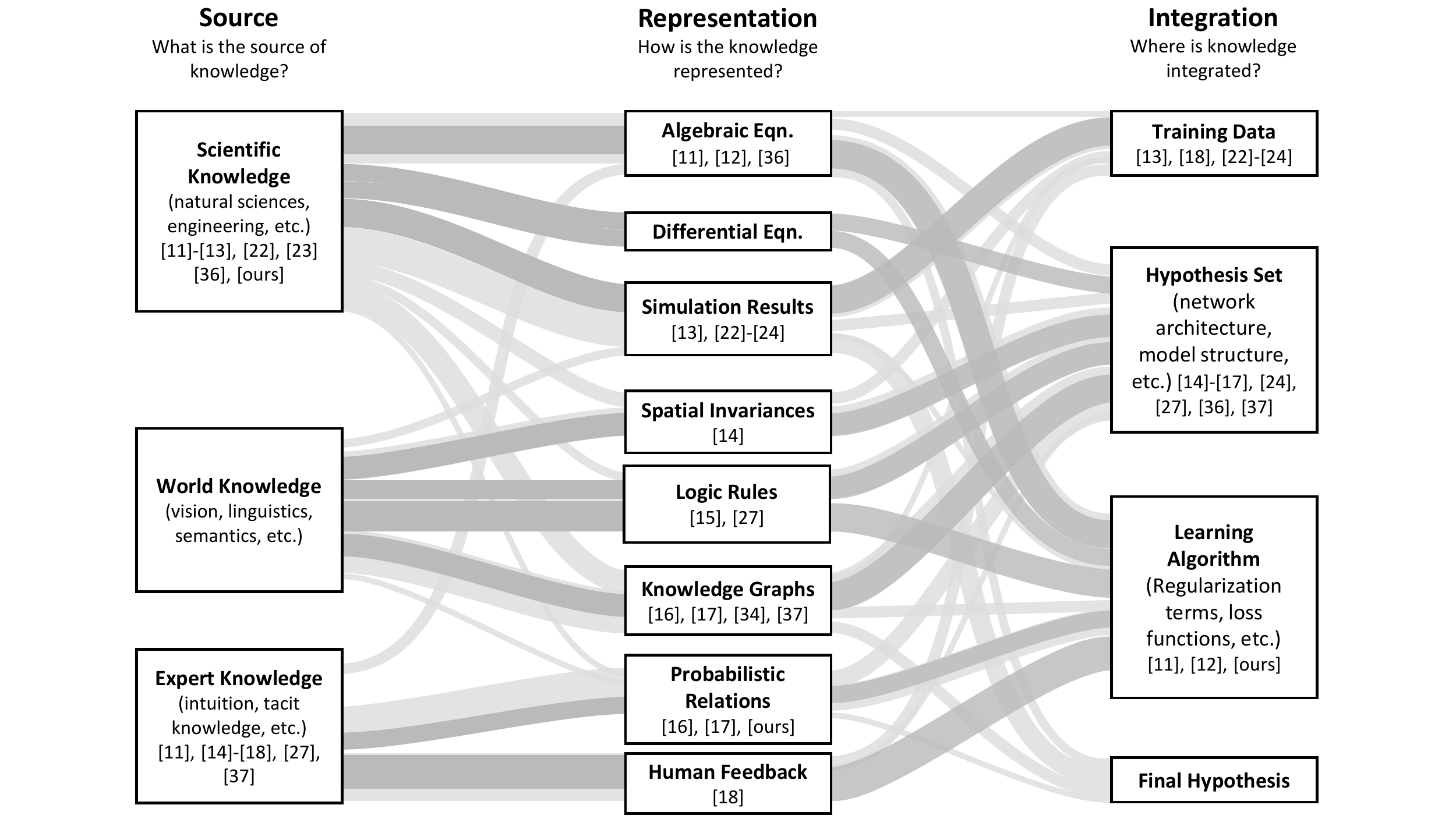}
  \caption{Knowledge informed machine learning can take many paths through the source, representation, and integration of knowledge. Papers that are knowledge informed machine learning, related to PHM, and discussed in this research, are highlighted, along with our work. (image modified from von Rueden et al. \cite{von2019informed} and used under CC by 4.0)}
  \label{fig:sankey}
\end{figure}

In their survey paper, \citeauthor{von2019informed} \cite{von2019informed} provide an articulate taxonomy for knowledge informed machine learning. They segment the different approaches within the field by the source of the knowledge; the method of representing that knowledge; and where the knowledge is integrated into the machine learning process. Figure~\ref{fig:sankey}, below, shows various paths, through the source, representation, and integration, by which knowledge informed machine learning can be applied. The articles referenced in this paper, and pertaining to PHM and knowledge informed machine learning, are highlighted.

\begin{figure}
  \centering
  \includegraphics[width=1.0\linewidth]{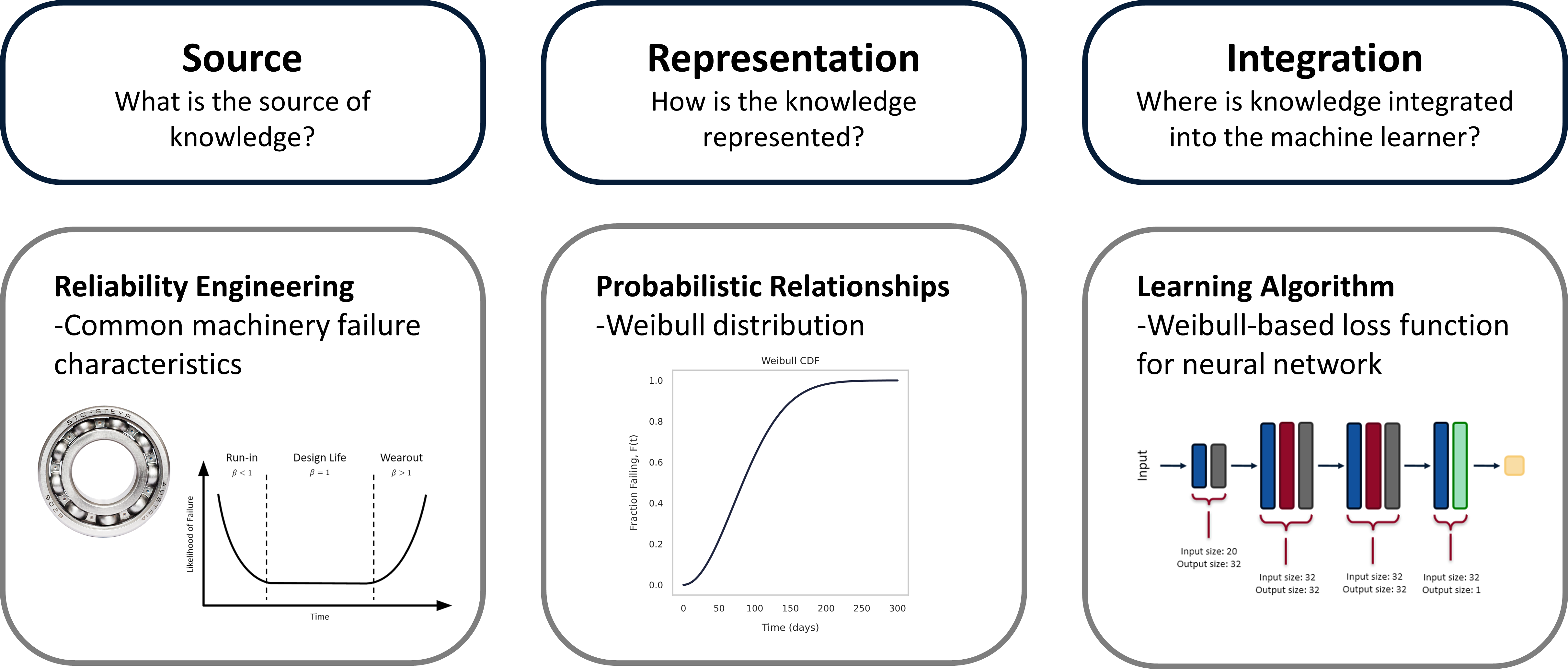}
  \caption{The taxonomy of knowledge informed machine learning (source, representation, and integration) and how it relates to the present study. In this experiment, knowledge is sourced from reliability engineering. Reliability engineering often involves the careful collection of machinery failure data, like that of a bearing. Through careful collection of the data, the mechanics of the failures can be understood, as in the “bathtub” curve. The knowledge from reliability engineering is then represented through the Weibull distribution, and in particular the Weibull cumulative distribution function (CDF). Finally, the knowledge is integrated into a neural network, during training, through a Weibull-based loss function. }
  \label{fig:source}
\end{figure}

The goal of this research is to demonstrate a new method for embedding prior knowledge in a remaining-useful-life (RUL) task with the use of a knowledge-based loss function. To that end, and as shown in Figure~\ref{fig:source}, the knowledge source derives from the field of reliability engineering with the common understanding of how machinery fail, which is represented through the probabilistic relationship of the Weibull distribution. Finally, the knowledge is integrated into the learning algorithm of a neural network through a Weibull-based loss function.

Knowledge informed machine learning was used in this research to predict remaining useful life (RUL) of bearings, using the IMS \cite{lee2007bearing} and PRONOSTIA \cite{nectoux2012pronostia} bearing data sets. As shown in Figure~\ref{fig:source}, the knowledge source is derived from the field of reliability engineering, with the common understanding of how machinery fail. The knowledge was represented through the probabilistic relationship of the Weibull distribution, and in particular the Weibull cumulative distribution function (CDF). The knowledge was integrated into the learning algorithm of a neural network through a Weibull-based loss function. 

The use of the Weibull-based loss function produced statistically significant results and was shown to outperform most models using traditional loss functions. The method excelled on the PRONOSTIA data set. The PRONOSTIA data is seen as less informationally rich, and thus, the machine learner relied on the Weibull-based loss functions more heavily. The Weibull-based loss functions demonstrated mild effectiveness on the IMS data. The IMS data set is seen as more realistic, but is also smaller in size, which contributed to the marginal performance of the method. Further limitations, and a discussion of results, is held in Section \ref{results}.

Although the research does not present a state-of-the-art feature engineering or neural network approach, the methods within are straightforward and can be readily implemented on other RUL and PHM tasks already using neural networks. Ideally, the method would be combined with regularization techniques, advanced feature engineering, and the tuning of model parameters, to achieve the best model for a particular task and data set. To that end, all the code and analysis has been made publicly available, with extensive documentation, to assist other researchers in their endeavors. Readers are also encouraged to visit one of the numerous surveys on RUL and PHM to gain a deeper understanding of classical and modern approaches \cite{kothamasu2006system, kordestani2019failure, lee2014prognostics, sikorska2011prognostic, lei2018machinery, rezamand2020critical}.

Overall, the goal of the research is to demonstrate the benefit of a knowledge informed machine learning technique in a PHM setting, through the novel Weibull-based loss function, and to help readers conceptually understand knowledge informed machine learning. To the best of our knowledge, there have been no prior papers that provide an overview of knowledge informed learning, from a PHM perspective, as defined by von Rueden. Our overview is not exhaustive but will provide the reader with a foundation to perform further investigation.

The following introductory sections expand on the paths within knowledge informed machine learning with a focus on how they are applied in PHM settings. 

\subsection{Source of Knowledge} \label{knowledge_source}
There are three primary sources to leverage in knowledge informed machine learning; they are scientific knowledge, world knowledge, and expert knowledge. Of these, scientific knowledge is most pertinent to PHM. Expert knowledge is also a source relevant to PHM and has ample room for further exploitation. World knowledge, however, is not as relevant to PHM research and will not be covered here.

Scientific knowledge, in the context of knowledge informed machine learning, can derive from a wide range of fields and engineering disciplines. Within PHM, Wang et al. use an empirical equation, sourced from the tool wear discipline, to improve their neural network used for wear prediction \cite{wang2020physics}. Yucesan et al. leverage the L10 bearing life equations, derived from mechanical engineering, to augment their wind turbine bearing failure model \cite{yucesan2019wind}. Sobie et al. use high resolution bearing dynamic simulations as a method of training their neural network, illustrating another approach of harnessing scientific knowledge in machine learning \cite{sobie2018simulation}.

Expert knowledge can also be a source of information to integrate into machine learning. Expert knowledge is less formal than scientific knowledge and can exist as intuition or tacit knowledge. Within PHM, expert knowledge was codified through a monotonicity constraint (explained further in Section~\ref{knowledge_based_loss_function}) for tool wear prediction \cite{wang2020physics}. \citeauthor{baseman2018physics} use expert knowledge in their research to predict failures in high-performance-computing memory. Specifically, they use the knowledge of spatial interdependence between memory devices \cite{baseman2018physics}.

\citeauthor{kordestani2019failure}, in their recent survey of failure prognosis applications \cite{kordestani2019failure}, discuss the limited use of expert knowledge by other PHM researchers. We too note less PHM research that explicitly leverages expert knowledge as defined by \citeauthor{von2019informed} However, expert knowledge can also be used in PHM applications through fuzzy networks \cite{vassilopoulos2008adaptive}, Bayesian networks \cite{kallen2005optimal, weidl2005applications}, and active learning \cite{nguyen2015active}.

The source of knowledge, in this research, comes largely from the field of reliability engineering (scientific knowledge). The discipline, described below, has codified practices and methods for improving equipment life and preventing failures, which can be used in a knowledge informed machine learning framework.

\subsubsection{Reliability Engineering} \label{reliability_engineering}
Reliability engineering is the process of ``preventing, assessing, and managing failures'' \cite{kapur2014reliability}. Today, reliability engineering is practiced across a broad range of industries and is relevant wherever PHM is employed. A reliability engineer uses tools from a range of fields, from statistics to continuous improvement, to prevent and predict failures.

\begin{figure}
  \centering
  \includegraphics[width=0.6\linewidth]{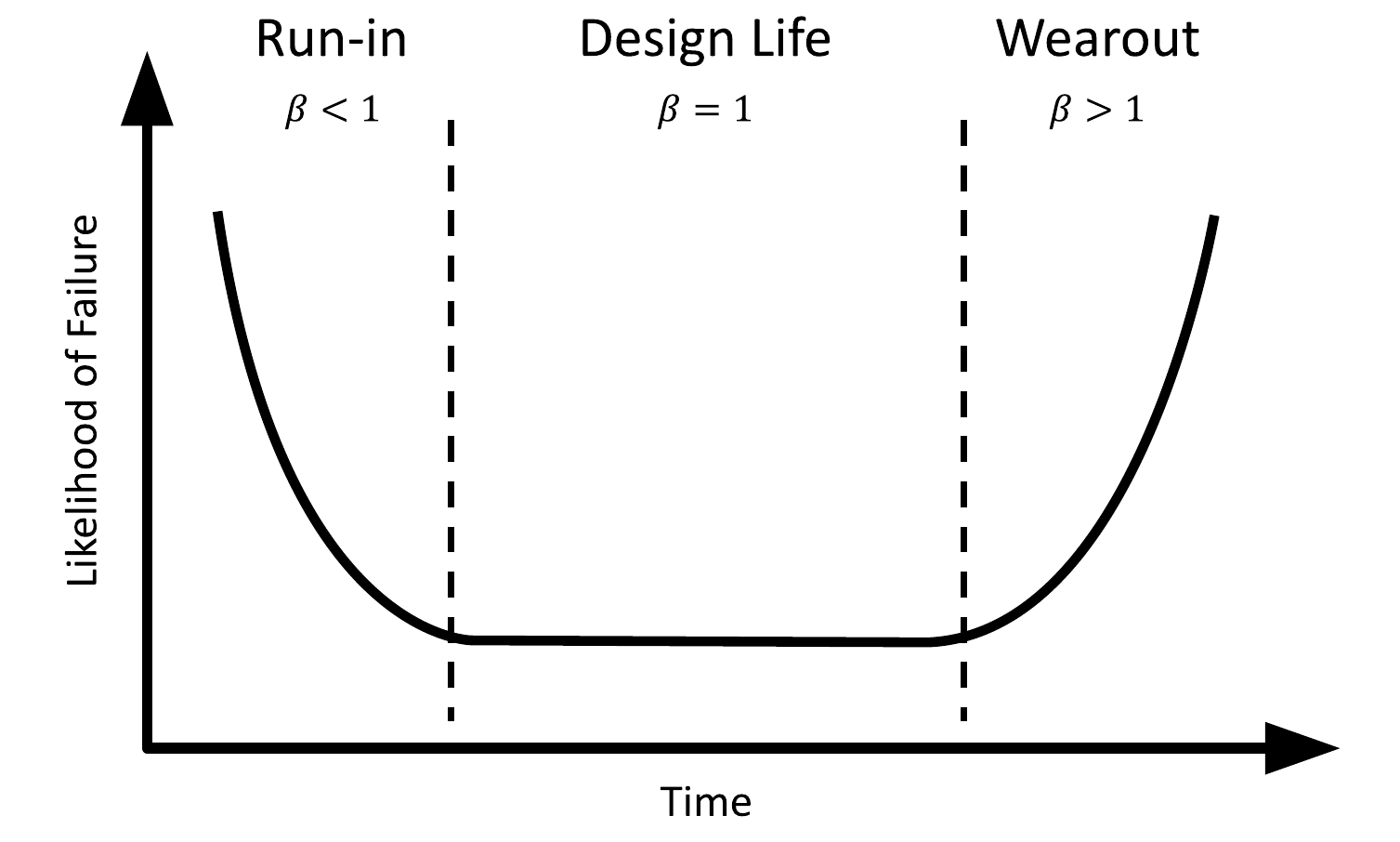}
  \caption{The ``bathtub'' curve demonstrates the likelihood of a component failing over its life. During each period or run-in, design life, and wearout, there is $\beta$ value that corresponds to the shape factor of its Weibull distribution.}
  \label{fig:bathtub}
\end{figure}

The understanding of how, and why, components fail is a regular undertaking in reliability engineering. Common mechanical components, such as bearings, generally fail following a ``bathtub curve'', as illustrated in Figure~\ref{fig:bathtub}, that describes the likelihood of failure as a function of time in use. 

During the initial period (typically referred to as ``run-in''), a bearing will have a higher likelihood of failure due to introduced problems such as manufacturing defects or installation errors. Once the bearing has passed its run-in stage it enters a prolonged period of its design life, where the likelihood of failure is relatively low and constant. Finally, the bearing enters a stage of wearout when the bearing has an increasing likelihood of failure.

Collecting failure data of components in operation, in an industrial environment or elsewhere, is common practice in reliability engineering. Importantly, the failure data is diligently collated with a deep understanding of failure mechanics and failure modes \cite{ja10111999evaluation, international2008iaea}. The failure data can then be used to parametrize a Weibull distribution, as discussed in Section~\ref{representation_knowledge}.

\subsection{Representation of Knowledge} \label{representation_knowledge}
Knowledge can be represented through a variety of methods, as illustrated in Figure~\ref{fig:sankey}. Knowledge represented through algebraic equations is one common method seen. As noted in Section~\ref{knowledge_source}, Wang \cite{wang2020physics} and Yucesan \cite{yucesan2019wind} both use algebraic equations for their PHM application of detecting tool wear and bearing failures, respectively.

Physics-based models are often seen in PHM applications. Additionally, these physics-based models can be used to represent knowledge in a knowledge informed machine learning context. \citeauthor{sobie2018simulation} generated their data from a three-degree-of-freedom model of bearing dynamics \cite{sobie2018simulation}. Similarly, \citeauthor{berri2019real} used a high-fidelity simulation to build their machine learning model for aircraft actuator RUL prediction \cite{berri2019real}. \citeauthor{abbas2007intelligent} used standard simulation software, found in the automotive industry, to generate data and build a prognostics system for detecting faults in car electrical systems \cite{abbas2007intelligent}. Finally, \citeauthor{liao2016hybrid} used a lumped parameter model, a physics based model, of a lithium ion battery to enhance their machine learner’s prediction of battery RUL \cite{liao2016hybrid}.

Human feedback can be an effective method of representing expert knowledge in knowledge informed ML. Active learning is a subset of human feedback whereby the machine learner selects data that is novel. The selected data is then labelled by a human expert \cite{settles2009active}. Thus, the active learning system can train its machine learner on less, and more information rich, data.

\citeauthor{nguyen2015active} \cite{nguyen2015active} use an active learning paradigm to create a partial discharge detection system, used on industrial electrical systems, which is trainable on much less data through active learning. Their method proves accurate with minimal degradation in performance. Active learning appears to be an underexplored modality within PHM. The technique shifts the development focus from a model-centric to a data-centric approach \cite{ng_chat_nodate}, which can be beneficial in real-world PHM applications where data is abundant but ``messy''. %%% NEED TO FIX THIS CITATION

Fuzzy neural networks, and the logic rules they employ, can be used to represent expert knowledge as well. Human knowledge is embedded through a series of “if-then” rules, which often derives from the insights of a domain expert. \citeauthor{vassilopoulos2008adaptive} \cite{vassilopoulos2008adaptive} employ an adaptive neuro-fuzzy inference system (ANFIS) to model fatigue behaviour in composites and predict RUL. \citeauthor{wang2004prognosis} \cite{wang2004prognosis} use a fuzzy neural network to provide gear wear forecasting.

A Bayesian network is a method of representing expert knowledge through probabilistic relations. When developing a Bayesian network, experts may be involved in defining the variables, establishing conditional relationships, or setting the conditional probabilities. \citeauthor{kallen2005optimal} \cite{kallen2005optimal} enhance the risk-based-inspection process for industrial pressure vessels using a gamma process and a Bayesian update model. The use of the Bayesian model allows experts to account for imperfect inspection results. \citeauthor{weidl2005applications} \cite{weidl2005applications} provided insight into causes of abnormalities in a complex industrial process using a Bayesian network. Deep expert knowledge of the process, and a firm understanding of potential failure modes, was required to build their network structure.

For the purposes of this research, the knowledge is represented through the probabilistic relationship of the Weibull distribution, and in particular the Weibull cumulative distribution function (CDF), described in the section below. The probabilistic relationship exists a priori from
reliability engineering and failure data. Others use probabilistic relationships for RUL prediction \cite{kim2012bearing}, but often they generate the probability as a part of a feature engineer process or machine learning pipeline.

\subsubsection{Weibull Distribution}
The likelihood of a component failing can be well represented by the Weibull distribution, which was first formalized by Walloddi Weibull in his seminal 1951 paper \cite{weibull1951statistical}. Today, the Weibull distribution is widely used in reliability engineering due to its expressiveness and ability to model a wide range of failure types \cite{martinsson2016wtte}. In addition, there are multiple examples of the Weibull distribution being used in PHM research across a variety of applications \cite{guo2009reliability, sutherland2003prognostics, goode2000plant}.

\begin{figure}
  \centering
  \includegraphics[width=0.9\linewidth]{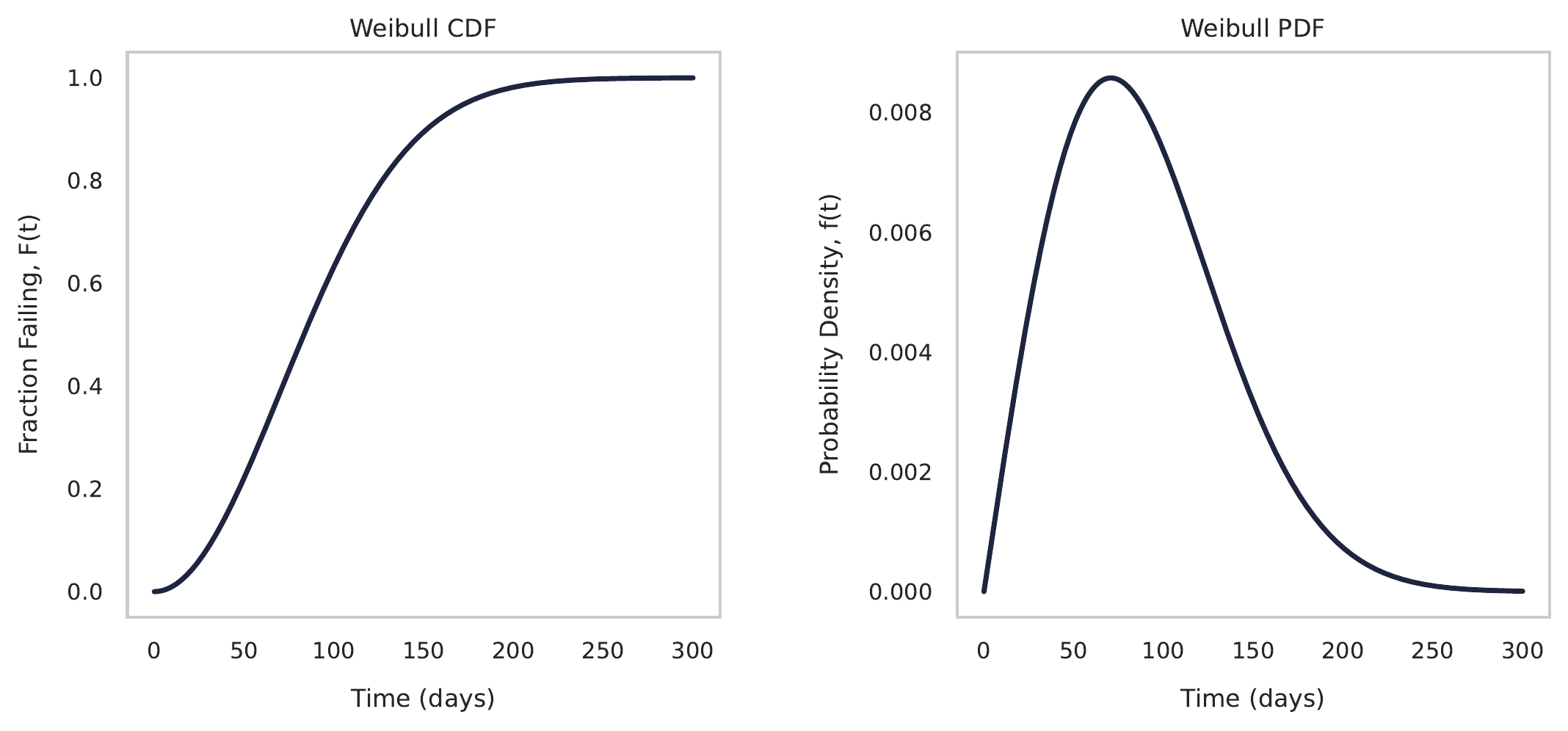}
  \caption{A Weibull CDF and its corresponding PDF. $\beta = 2.0$, $\eta = 100$ days}
  \label{fig:weibull_cdf}
\end{figure}

The two parameter Weibull cumulative distribution function (CDF) is the probability of failure, $F(t)$, given a time, $t$, as shown in Figure~\ref{fig:weibull_cdf} Alternatively, it is the fraction of units failing up-to time $t$. The CDF is used in this research.

The CDF function is defined as:
    
    \begin{equation}
        F(t) = 1 - e^{-(t/\eta)^\beta}
        \label{eqn:cdf}
    \end{equation}
    
\noindent where $\beta$ is the shape parameter and $\eta$ is the characteristic life. When plotted on a log-log scale $\beta$ is the slope of the line. The characteristic life, $\eta$, is the age where 63.2\% of the units have failed.

The shape parameter, $\beta$, corresponds to where a component commonly fails at the end of its useful life. As illustrated in Figure~\ref{fig:bathtub}, a component that has a high probability of failure in its run-in period (infant mortality) will have a $\beta$ value of less-than one. A component that has a random probability of failure during its design life will have a $\beta$ value of approximately one. Finally, a component that has a high probability of failure due to wearout will have a $\beta$ value greater than one.

A reliability engineer’s job may entail creating a library of shape parameters ($\beta$) for common and critical components. The shape parameters are created through diligent analysis of failures and a thorough collection of history. With enough failures, a characteristic life ($\eta$) can be estimated. This knowledge, of the $\beta$ and $\eta$, exists outside of the machine learning pipeline, and can be utilized to parametrize a Weibull distribution.

\subsubsection{Using the Weibull Distribution with Insufficient Failure Data}
In the ideal case, as described above, the shape parameter ($\beta$) and characteristic life ($\eta$) will be produced from the reliability engineering process through a rich history and careful analysis of failures. The method of Reliability Centered Maintenance (RCM) codifies this practice and is commonplace in industry \cite{ja10111999evaluation, international2008iaea}. However, in a newer manufacturing facility, for example, there will not be sufficient failure history to properly fit a Weibull distribution on components of concern. In this situation, where only a few examples of failures exist, the following equation can be used to estimate the characteristic life given a reasonable estimate for $\beta$:

\begin{equation}
        \eta = \left[ \sum_{i=1} \frac{t_i^\beta}{r}\right]^{1/\beta} 
        \label{eqn:weibays}
\end{equation}
    
where:

$t =$ time or cycles, \newline
$r =$ number of failed units, \newline
$N =$ number of failures plus incomplete run-to-failure samples

Equation~\ref{eqn:weibays} is derived using the maximum-likelihood method. With the estimated values of $\beta$ and $\eta$, a Weibull distribution can be drawn. This above procedure is sometimes called the Weibayes method \cite{abernethy2004new}. 

In the case of the newer manufacturing facility, the estimates of the shape parameter ($\beta$) may be derived from a sister manufacturing facility that has a similar layout, but with a longer operational history.

We conducted simple RUL experiments on two bearing data sets in this research. Both bearing data sets have insufficient failure data to properly parametrize a Weibull distribution. As such, we used the Weibayes equation and some generally acceptable values for bearing shape parameters \cite{abernethy2004new}, as shown below in Table~\ref{tab:bearing_estimates}. Further limitations to this approach are discussed in Section~\ref{limitations}.

\begin{table}[th]
% \tiny
  \caption[$\beta$ estimates for bearings]{$\beta$ estimates for bearings}
  \centering
  \begin{tabular}{p{0.25\linewidth}p{0.15\linewidth}}
    \toprule

    Component Type  & $\beta$ Estimate\\
    \midrule
    
    Ball bearing   & 2.0\\

    Roller bearing & 1.5\\
    \bottomrule
  \end{tabular}
\label{tab:bearing_estimates}
\end{table}

\subsection{Integration of Knowledge}
Knowledge integration is the final step in knowledge informed machine learning. There are four primary methods of knowledge integration: using training data, using hypothesis sets, using learning algorithms, and using a final hypothesis check.

Simulations, as shown by Sobie \cite{sobie2018simulation} and Berri \cite{berri2019real}, are often integrated into the training data in knowledge informed machine learning. Creating the simulations is explicit knowledge that lives outside the domain of machine learning. The simulations can then be used to produce data that is used to train a machine learner.

Wang \cite{wang2020physics} and Yucesan \cite{yucesan2019wind} use their empirically based algebraic equations as an additional source of information for their training data. In essence, they use the information from their algebraic equations and fuse it with the data collected via sensors -- a process called data fusion \cite{zheng2015methodologies}. Subjectively, there are many empirical equations across various engineering disciplines that could be integrated into the training data in a similar way.

Pre-existing knowledge can also be used to constrain the space over which a machine learner searches; that is, knowledge is integrated into the hypothesis set. \citeauthor{lu2017physics} \cite{lu2017physics} apply this method of integrating knowledge into the hypothesis set in their application of electrochemical micro-machining. They embedded several linear equations, approximated with Taylor functions, as linear activations in the neural network. The network was then used to assess product quality. 

The use of Bayesian \cite{kallen2005optimal, weidl2005applications} and fuzzy networks \cite{vassilopoulos2008adaptive, wang2004prognosis} is also an example of knowledge integrated into the hypothesis set. As noted in Section~\ref{knowledge_source}, \citeauthor{baseman2018physics} \cite{baseman2018physics} leveraged spatial interdependence to detect failures in computing memory. The spatial interdependence is expressed through a graph representation, also called a graph network, of the data. The use of a graph network is an effective method of integrating knowledge, often from experts, through a hypothesis set. The method is also used by \citeauthor{ruiz2021system} \cite{ruiz2021system} to represent interdependence between equipment in a chemical facility. They use the graph representation, and a convolutional neural network, to assess the health of a chlorine dioxide generation system.

Integration of knowledge through a final hypothesis often involves a consistency check \cite{von2019informed}. That is, the results from a machine learning model are checked against external knowledge, which can be used to measure the performance of the model. To the authors’ awareness, this method is not common in PHM.

Finally, knowledge can be integrated in a machine learner via a learning algorithm. A common approach is to integrate the external knowledge into the loss function. The loss function is then used in the training of a neural network. The knowledge-based loss function acts as a constraint on the machine learner during its training. This method, and specifically, a Weibull-based loss function, is used in this research and described below.

\subsubsection{Knowledge-based Loss Functions} \label{knowledge_based_loss_function}
The training of a neural network is achieved by minimizing a loss function, in a process called back propagation \cite{rumelhart1985learning}. A common loss function is the mean-squared-error (MSE) loss, as shown by:

\begin{equation}
    \Lagr_\text{MSE} = \frac{1}{n}\sum_{i=1}^{n}(y_i - \hat{y_i})^2
\end{equation}

\noindent where $y$ is the true label and $\hat{y}$ is the predicted label. 

A knowledge-based loss functions, also called a hybrid loss functions, combines domain specific knowledge with traditional label-based loss , such as the MSE loss. The knowledge-based loss function can be represented by:

\begin{equation}
    \Lagr_\text{hybrid} = \Lagr_\text{MSE} + \lambda \Lagr_\text{domain}
\end{equation}

\noindent where $\Lagr_\text{MSE}$ is the mean-squared-error loss function, $\Lagr_\text{domain}$ is the domain specific loss function, and $\lambda$ is a hyper-parameter to set the weight of the domain loss term.

There is a growing library of knowledge-based loss functions and methods being used in today’s machine learning literature. \citeauthor{stewart2017label} \cite{stewart2017label} used basic equations of motion to help train a neural network to predict the trajectory of a thrown object. Importantly, they employed a label-free method that did not require ground-truth labels during training. 

\citeauthor{wang2020physics} \cite{wang2020physics} used the fact that tool wear tends to increase, monotonically, to help guide the training of their neural network. This method, the monotonicity constraint, is embedded in the loss function.

Another method of embedding knowledge in the loss function is through an approximation constraint. An approximation constraint enforces a realistic upper and lower bound on the results of a neural network during training. The bounds can be defined by expertise, a probabilistic relation, or else wise. If the results, during training, deviates from the realistic bounds the approximation constraint will apply a corrective force on the neural network. 

\citeauthor{muralidhar2018incorporating} \cite{muralidhar2018incorporating} demonstrate the effectiveness of the approximation constraint on synthetic data. \citeauthor{stewart2017label} \cite{stewart2017label} also use the approximation constraint to predict the position of a walking person. They assumed that a person walks at a constant velocity over short periods of time, which they then integrate into their loss function. 

\subsubsection{Weibull-Based Loss Function}
The Weibull cumulative distribution function, $F(t)$, can be combined as an approximation constraint loss function. The Weibull-based loss function is as follows:

\begin{equation}
    \Lagr(\bm{t},\bm{\hat{t}}) = \underbrace{\frac{1}{n}\sum_{i=1}^{n}( t_i - \hat{t_i})^2}_\text{MSE loss} + \underbrace{\lambda \frac{1}{n}\sum_{i=1}^{n}( F(t_i) - F(\hat{t_i}))^2}_\text{Weibull loss}
\end{equation}

\noindent where $n$ is the number of samples in the data set. $F(t_i)$ is the true fraction failing at time $i$, which is calculated with the CDF function as shown in Equation~\ref{eqn:cdf}. $F(\hat{t_i})$ is the estimated fraction failing at time $i$. The estimated fraction failing is calculated from the output of neural network after each training step.

The above Weibull-based loss function can be used both with sufficient or insufficient failure data. When insufficient failure data exists an estimate of the shape parameter ($\beta$) must be used, in combination with the Weibayes equation, to generate an estimated characteristic life ($\eta$). The experiment in this work uses the latter approach with the Weibayes equation. 

%%%%%%%%%%%%%%%%%%%%%%%%%%%%%%%%%%%%
%%%%%%%%%%%%%%%%%%%%%%%%%%%%%%%%%%%%
%%%%%%%%%%%%%%%%%%%%%%%%%%%%%%%%%%%%
%%%%%%%%%%%%%%%%%%%%%%%%%%%%%%%%%%%%

\section{Data and Model}
\subsection{Data Description}
The Weibull-based loss function was tested on neural networks trained on the common IMS \cite{lee2007bearing} and PRONOSTIA \cite{nectoux2012pronostia} bearing data sets. Both data sets are available on the \href{https://ti.arc.nasa.gov/tech/dash/groups/pcoe/prognostic-data-repository/}{NASA Prognostics Data Repository}. Table~\ref{tab:data_attributes} details key parameters of these data sets.

\begin{table}[ht]
\centering
\caption{Key attributes of the IMS and PRONOSTIA bearing data sets}
\begin{tabular}[t]{lcc}
\toprule
&IMS&PRONOSTIA\\
\midrule
Sampling Frequency &20,480 Hz&25,600 Hz\\
Operating Speed&2000 RPM&1800, 1650, 1500 RPM\\
Static Loading&26.7 kN& 4, 4.2, 5 kN\\
Bore Diameter&49.2 mm& 20 mm\\
Max Runtime & 34 days 12h&126 min\\
\bottomrule
\end{tabular}
\label{tab:data_attributes}
\end{table}%

\begin{figure}
  \centering
  \includegraphics[width=0.5\linewidth]{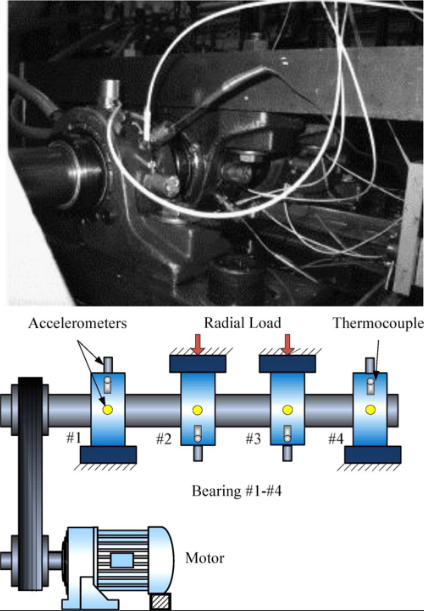}
  \caption{IMS layout \cite{gousseau2016analysis}}
  \label{fig:ims}
\end{figure}

The IMS bearing data set is from the University of Cincinnati, and consists of three run-to-failure experiments on a loaded shaft. The shaft, as shown in Figure~\ref{fig:ims}, was supported by four roller bearings, with the housing of each bearing instrumented with a vertical and horizontal accelerometer. 

The loading and operating speeds were constant between IMS experiments. The accelerometer data was collected approximately every 10 minutes in each run-to-failure data set, with each sample being one minute in length.

\begin{figure}
  \centering
  \includegraphics[width=0.8\linewidth]{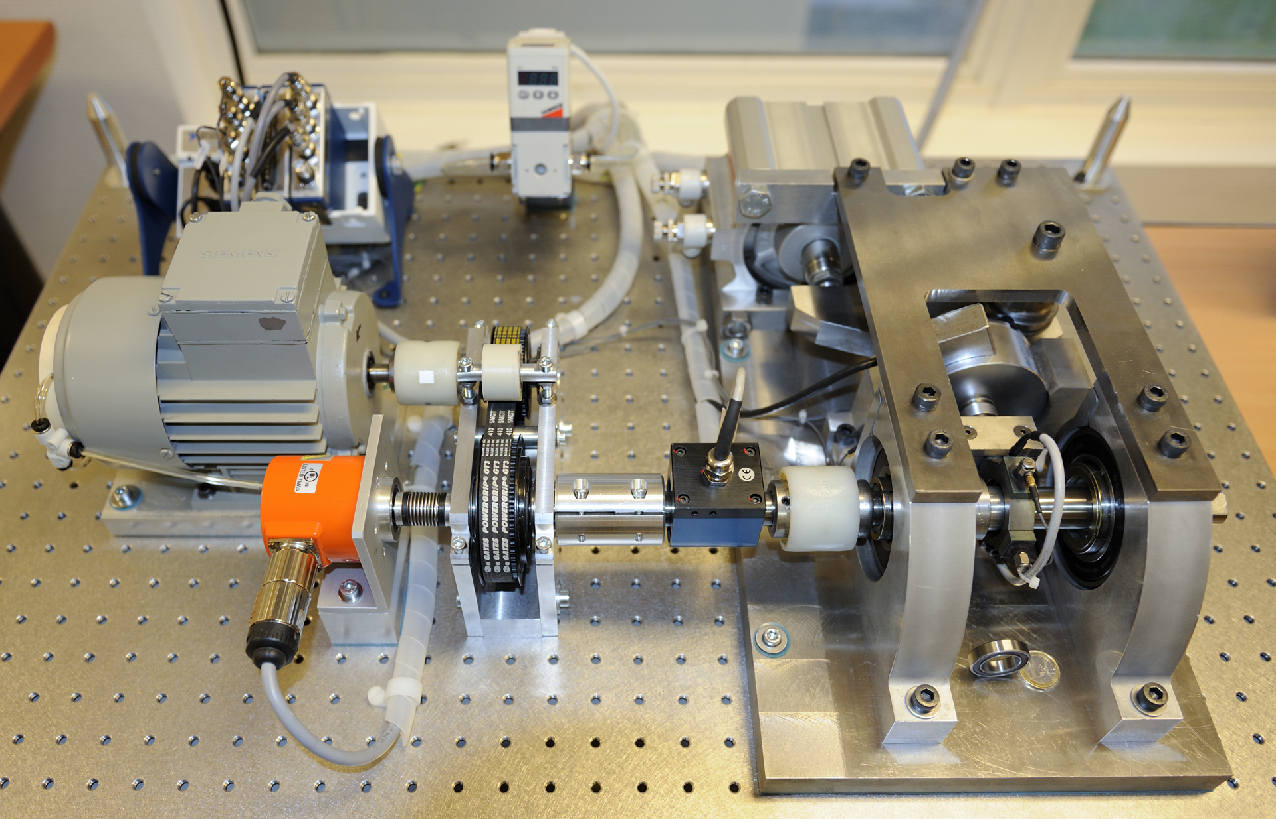}
  \caption{The PRONOSTIA experimental test-bench \cite{nectoux2012pronostia}}
  \label{fig:pronostia}
\end{figure}

The PRONOSTIA bearing data set, from the FEMTO-ST Institute in France, is a set of 17 accelerated run-to-failure experiments. Acceleration and temperature data was collected on an test-bench, as shown in Figure~\ref{fig:pronostia}, that exposed a bearing to variable loads and speeds. 

The PRONOSTIA experiments were conducted under three operating conditions as shown in Table~\ref{tab:data_attributes}. The acceleration data was collected every 10 seconds, with each sample being 1/10-second in length. Only the acceleration data was used in the research presented here.

\subsection{Preprocessing and Feature Engineering}
The purpose of this research is to understand if a Weibull-based loss function improves the results of a RUL estimation task. As such, minimal preprocessing and feature engineering was performed on the data sets

\begin{figure}
  \centering
  \includegraphics[width=0.9\linewidth]{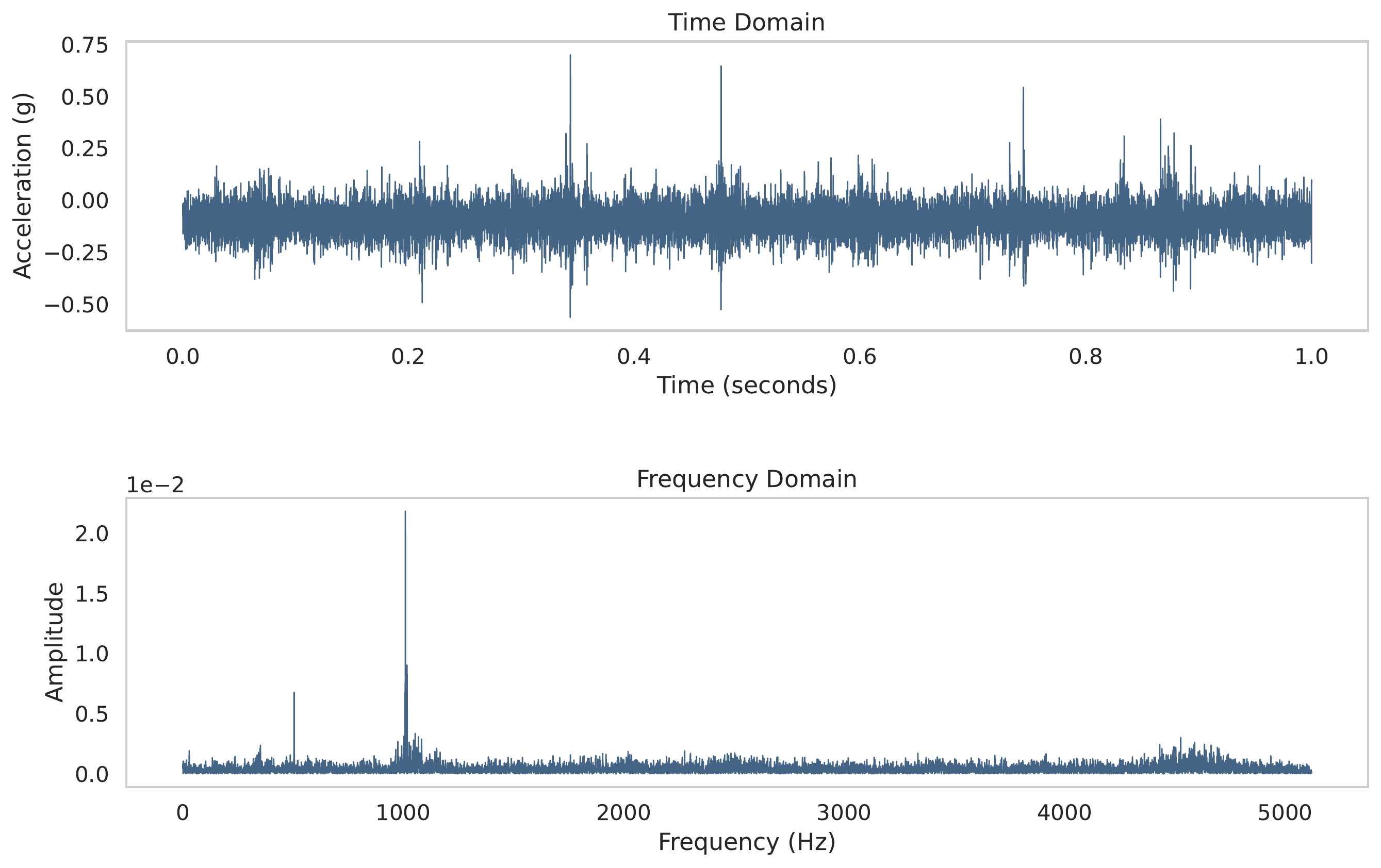}
  \caption{Time and frequency domain of an acceleration signal from the IMS data set}
  \label{fig:time_freq_ims}
\end{figure}

The IMS and PRONOSTIA data sets underwent similar preprocessing steps. First, each acceleration sample was detrended and then windowed with the Kaiser function. The samples were subsequently moved into the frequency domain with the fast-Fourier transform (FFT). Figure~\ref{fig:time_freq_ims} shows a representative time domain and frequency domain signal after preprocessing.

Finally, each FFT spectrum was segmented into 20 equally sized bins. The maximum amplitude within each bin became a value in the final vector, $x$, making a total of 20 values. As shown below, in Figure~\ref{fig:spectrogram}-a, each FFT spectrum can be plotted over time to form a spectrogram. Figure~\ref{fig:spectrogram}-b demonstrates the spectrogram after each FFT spectrum has been segmented into ``bins''. Each vertical ``slice'' of the spectrogram corresponds to one data vector, $x$, that would be used in a neural network. The Appendix contains the complete IMS and PRONOSTIA training, validation, and testing spectrograms, similar to that seen in Figure~\ref{fig:spectrogram}-b,

\begin{figure}
  \centering
  \includegraphics[width=1.0\linewidth]{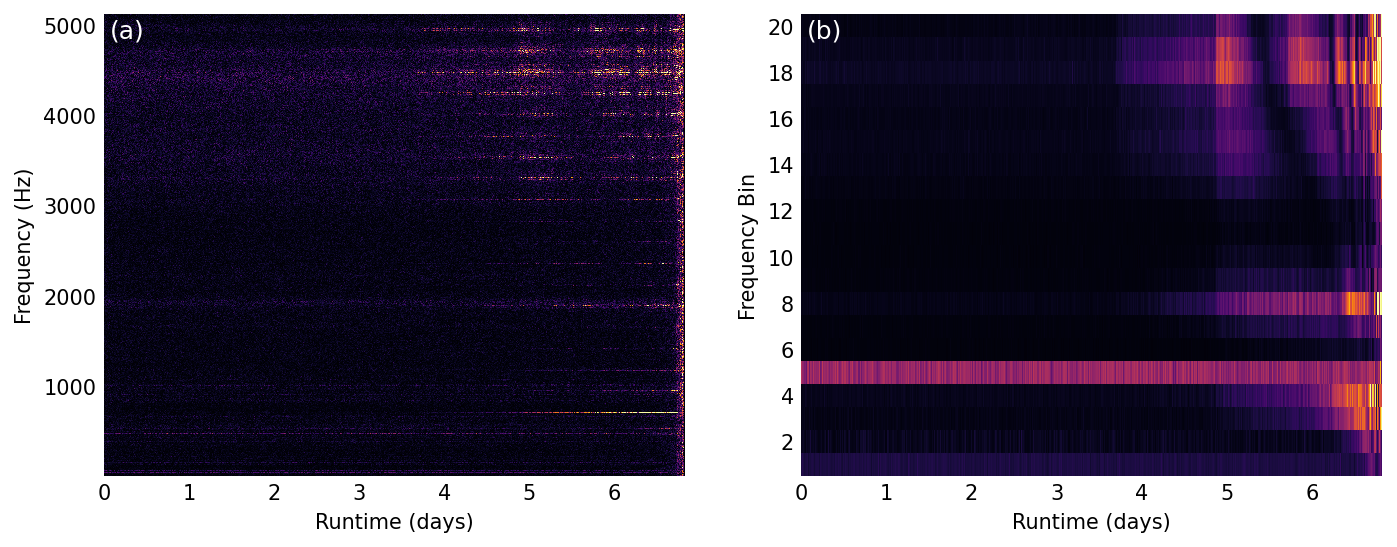}
  \caption{(a) Spectrogram from the second IMS experiment, bearing one, horizontal acceleration channel. (b) The resulting spectrogram after taking the maximum value in each of the 20 bins.}
  \label{fig:spectrogram}
 \end{figure}

Each acceleration sample has an associated date and time of its creation. The date and time can be used to represent the life percentage of the bearing, as done by \citeauthor{ali2015accurate} \cite{ali2015accurate}, given by:

\begin{equation}
    \text{Life Percentage} = \frac{t_i}{t_N} \times 100
\end{equation}

\noindent where $t_i$ is the run-time of the acceleration sample, $i$. $t_N$ is the total run-time of the bearing, and $N$ is the number of samples in the experiment. The life percentage became the label, $y$, for each $x$ vector.

\subsection{Model}
\begin{figure}
  \centering
  \includegraphics[width=0.8\linewidth]{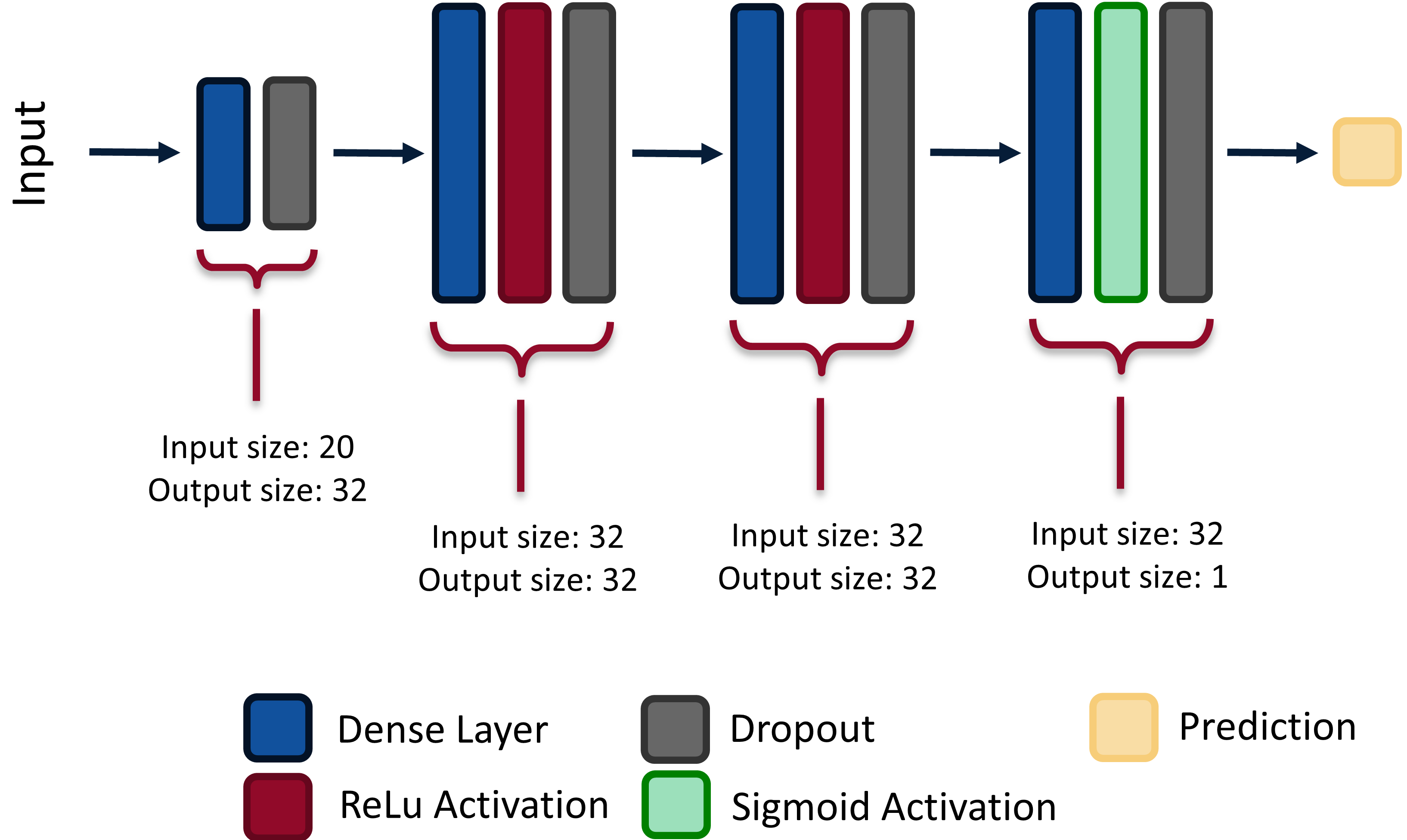}
  \caption{ One example of a neural network architecture tested during the experiment. The network takes a vector input of size 20 and has three hidden layers with 32 neurons in each layer. Each layer used dropout during training as a regularization method. The rectified linear unit (ReLu) activation was used throughout, except for the last layer which used a sigmoid activation.}
  \label{fig:dense_network}
 \end{figure}

Many feed-forward neural network architectures were tested, on both data sets, to evaluate the Weibull-based loss functions. The neural networks were built with differing number of layers and units per layer, as detailed in Table~\ref{tab:model_layers}. Figure~\ref{fig:dense_network} demonstrates what one of the models could look like.

\begin{table}[th]
% \tiny
  \caption{The number of layer and units per layer in the feed-forward neural networks}
  \centering
  \begin{tabular}{lc}
    \toprule

    Parameter  & Selection Choice\\
    \midrule
    Number of layers  & Between 2 and 7\\
    Number of units per layer &  16, 32, 64, 128, 256\\
    \bottomrule
  \end{tabular}
\label{tab:model_layers}
\end{table}

 For this research, the focus was on demonstrating the capability of the Weibull-based loss function without the need for complex feature engineering or neural network architecture. As such, the simple feed-forward neural network architecture was chosen. All models, even the largest model with 7 layers and 256 units per hidden layer, can be readily trained on a low-end GPU. Likewise, model prediction is fast, due to the simple model architecture, even without a GPU.
 
 Finally, nine loss functions, as shown in Table~\ref{tab:loss_functions}, were tested on each unique model architecture. The loss functions consisted of traditional loss functions, and Weibull-based loss functions. The Weibull-based loss functions can be further segmented into Weibull only loss functions (no traditional loss function included) and Weibull combined loss functions (the traditional loss function is included with the Weibull portion).

\begin{table}[th]
% \tiny
  \caption[Loss functions used]{Loss functions tested}
  \centering
  \begin{tabular}{p{0.25\linewidth}|p{0.5\linewidth}}
    \toprule

    Loss Function  & Equation\\
    \midrule
    
    MSE Loss ($\Lagr_\text{MSE}$)   & $\frac{1}{n}\sum_{i=1}^{n}(t_i - \hat{t_i})^2$\\
    \midrule
    
    RMSE Loss ($\Lagr_\text{RMSE}$)  & $\sqrt{\frac{1}{n}\sum_{i=1}^{n}(t_i - \hat{t_i})^2}$\\
    \midrule
 
    RMSLE Loss ($\Lagr_\text{RMSLE}$)  & $\sqrt{\frac{1}{n}\sum_{i=1}^{n}(\log{(t_i+1)} - \log{(\hat{t_i}+1)})^2}$\\
    \midrule 
    
    Weibull Only MSE Loss ($\Lagr_\text{Weibull-MSE}$)   & $\lambda \frac{1}{n}\sum_{i=1}^{n}(F(t_i) - F(\hat{t_i}))^2$\\
    \midrule
    
    Weibull Only RMSE Loss ($\Lagr_\text{Weibull-RMSE}$)  & $\lambda \sqrt{\frac{1}{n}\sum_{i=1}^{n}(F(t_i) - F(\hat{t_i}))^2}$\\
    \midrule
 
    Weibull Only RMSLE Loss ($\Lagr_\text{Weibull-RMSLE}$)  & $\lambda \sqrt{\frac{1}{n}\sum_{i=1}^{n}(\log{(F(t_i)+1)} - \log{(F(\hat{t_i})+1)})^2}$\\
    \midrule

    Weibull-MSE Combined Loss ($\Lagr_\text{Weibull-MSE-Comb}$)   & $\Lagr_\text{MSE} + \lambda  \Lagr_\text{Weibull-MSE}$\\
    \midrule
    
    Weibull-RMSE Loss ($\Lagr_\text{Weibull-RMSE-Comb}$)  & $\Lagr_\text{RMSE} + \lambda  \Lagr_\text{Weibull-RMSE}$\\
    \midrule
 
    Weibull-RMSLE Loss ($\Lagr_\text{Weibull-RMSLE-Comb}$)  & $\Lagr_\text{RMSLE} + \lambda  \Lagr_\text{Weibull-RMSLE}$\\

    \bottomrule
  \end{tabular}
\label{tab:loss_functions}
\end{table}

\section{Experiment} \label{experiment}

The data was split into two training, validation, and testing sets, as shown in Tables~\ref{tab:ims_data_splits} and \ref{tab:femto_data_splits}, using the horizontal acceleration values. Only the first six runs on the PRONOSTIA data set were used.

After the preprocessing and feature engineering steps, minimum/maximum scaling (between 0 and 1) was performed on the training sets. The same scaling was then applied to the respective validation and testing sets. The characteristic life, $\eta$, was calculated using the training data, the Weibayes equations, and a $\beta$ of 2.0, for each respective data set. Other shape parameters were briefly tried in prototyping. Anecdotally, a $\beta$ of 2.0 was found to provide the most stability during training. However, this is seen as a limitation to the experiment, further discussed in Section~\ref{limitations}, and should be explored further.

\begin{table}[th]
% \tiny
  \caption{IMS data splits. Only the horizontal channel used in each run.}
  \centering
  \begin{tabular}{ccc}
    \toprule

    Train  & Validation & Test\\
    \midrule
    run 2, bearing 1   & run 1, bearing 3 & run 1, bearing 4\\
    run 3, bearing 3 &  &\\
    \bottomrule
  \end{tabular}
\label{tab:ims_data_splits}
\end{table}

\begin{table}[th]
% \tiny
  \caption{PRONOSTIA data splits. Horizontal channel used in each run.}
  \centering
  \begin{tabular}{ccc}
    \toprule

    Train  & Validation & Test\\
    \midrule
    Bearing1\_1 & Bearing1\_2 & Bearing1\_3\\
    Bearing2\_1 & Bearing2\_2  & Bearing2\_3\\
    Bearing3\_1 & Bearing3\_2 & Bearing3\_3\\
    \bottomrule
  \end{tabular}
\label{tab:femto_data_splits}
\end{table}

A random search was used to select the hyperparameters for each model architecture. Table~\ref{tab:general_parameters} shows the parameters that were used in the random search. In general, a random search is more effective than a grid search for selecting hyperparameters \cite{bergstra2012random}.

%%%% HYPERPARAMETER TABLE %%%%%%%%%%
\begin{table}[th]
% \tiny
  \caption[Hyperparameters used in random search]{The hyperparameters used in the random search}
  \centering
  \begin{tabular}{p{0.25\linewidth}p{0.65\linewidth}}
    \toprule

    Parameter  & Selection Choice\\
    \midrule
    
    Batch size   & \texttt{32, 64, 128, 256, 512}\\
    \midrule
    
    Learning rate   & \texttt{0.1, 0.01, 0.001, 0.0001}\\
    \midrule
    
    Lambda   & Floating point number between 0 and 3  \\
    \midrule
    
    Number of layers    & Integer between 2 and 7\\
    \midrule

    Number of units per layer   & \texttt{16, 32, 64, 128, 256}\\
    \midrule
    
    Probability of dropout   & \texttt{0.1, 0.2, 0.25, 0.4, 0.5, 0.6}\\

    \bottomrule
  \end{tabular}
\label{tab:general_parameters}
\end{table}

\begin{figure}[h]
  \centering
  \includegraphics[width=1.0\linewidth]{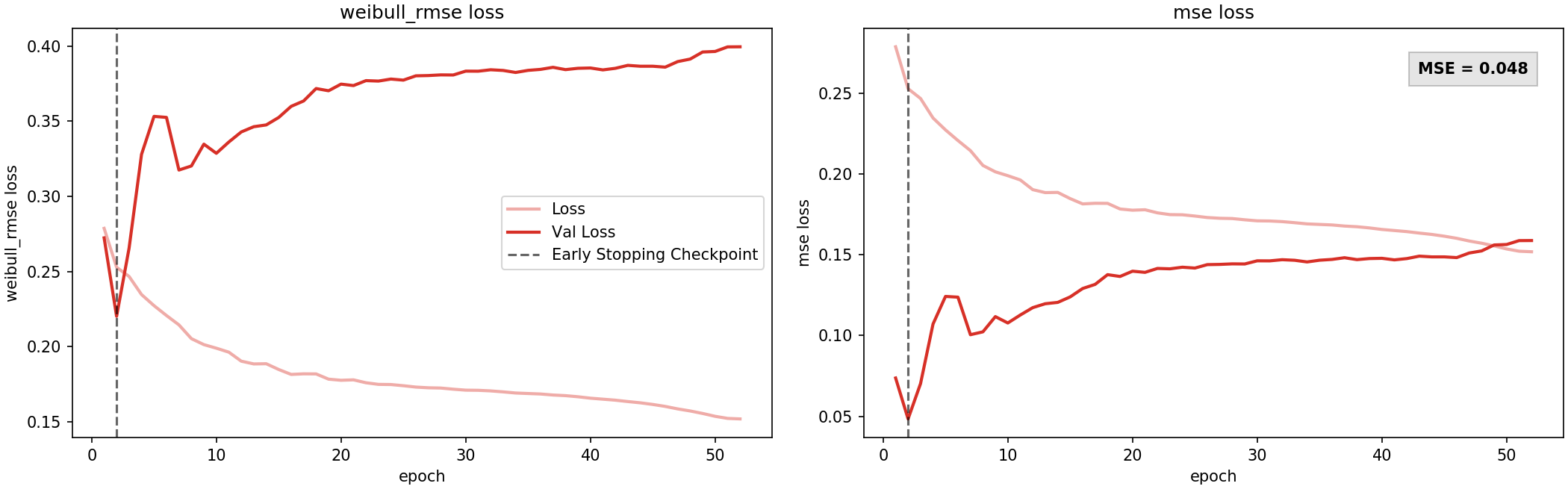}
  \caption{IMS learning curves for the top performing model. Early stopping was triggered at epoch 2, which became the final model.}
  \label{fig:ims_learning_curves}
 \end{figure}

\begin{figure}[h]
  \centering
  \includegraphics[width=1.0\linewidth]{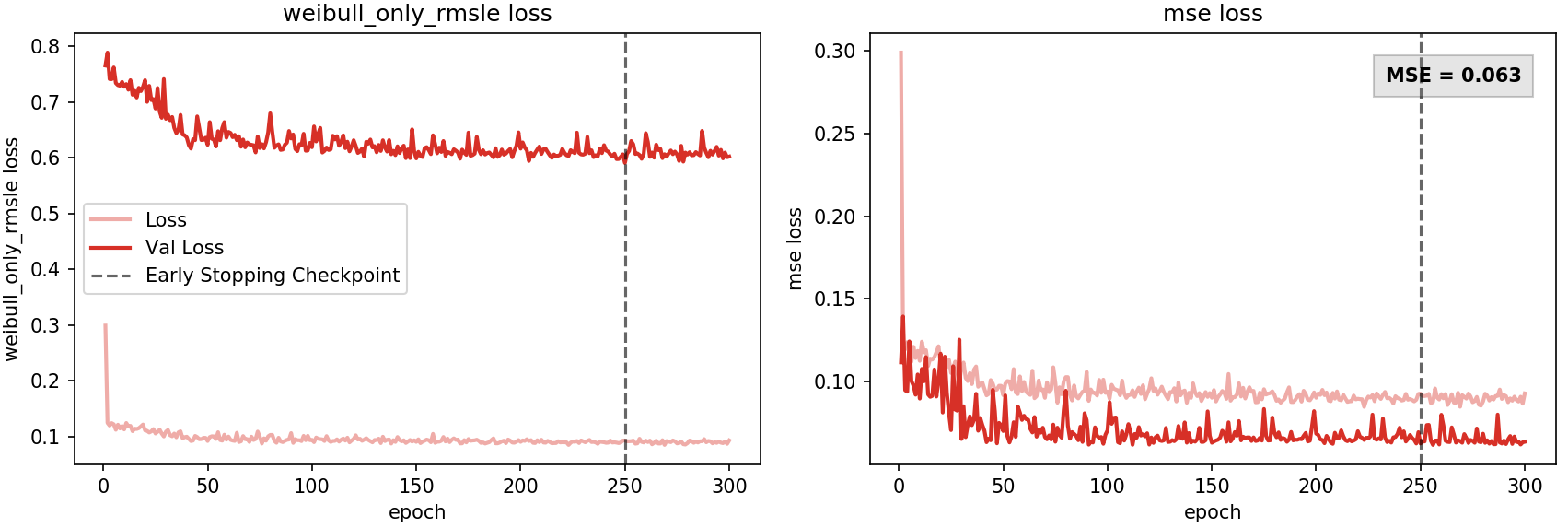}
  \caption{PRONOSTIA learning curves for the top performing model. Early stopping was triggered at epoch 250, which became the final model.}
  \label{fig:femto_learning_curves}
 \end{figure}

During training, the ADAM optimizer was used to manage learning rate \cite{kingma2014adam}. Early stopping, based on the validation loss, was used to prevent overfitting.  As an example of the early stopping, Figures~\ref{fig:ims_learning_curves} and \ref{fig:femto_learning_curves} show where the top performing models quit training.

The mean squared error (MSE), root mean squared error (RMSE), root mean squared log error (RSMLE), and the coefficient of determination ($R^2$) metrics were recorded for each model. $R^2$ is a metric that shows how well data fits to a regression line and is often seen in RUL experiments for measuring performance. $R^2$ was used as the primary evaluation metric.

Numerous model architectures were created during the random search. For the IMS data set, approximately 1000 architectures were tested. Each unique model architecture was used to train nine models, one for each loss function as shown in Table~\ref{tab:loss_functions}, totaling 9,375 models. A similar number of architectures were tested using the PRONOSTIA data set, totaling 8,495 models.

Finally, the data preprocessing was conducted in Python and the neural networks were constructed with PyTorch. The random search and model training was performed on multiple NVIDIA P100 GPUs using approximately 8 days of combined GPU time. The reader can review all the code used in the experiment on the author’s GitHub page\footnote[1]{\href{https://github.com/tvhahn/weibull-knowledge-informed-ml}{https://github.com/tvhahn/weibull-knowledge-informed-ml}}.

\section{Results} \label{results}
A threshold, based on the RMSE and $R^2$, was used to filter out poorly performing models from the random search. To be classified as a well performing model, a trained neural network had to achieve a $R^2$ greater than 0.2 and a RMSE less than 0.35 on each of the training, validation, and testing sets, as shown in Table~\ref{tab:thresholds_filter}. The thresholds were manually selected to be not overly restrictive, but effective enough to remove erroneous models.

\begin{table}[htp]
\centering
\caption{The threshold values used to filter out poorly performing models.}
\begin{tabular}[t]{lccc}
\toprule
&Train&Val&Test\\
\midrule
$R^2$ &> 0.2&> 0.2&> 0.2\\
RMSE &< 0.35&< 0.35&< 0.35\\

\bottomrule
\end{tabular}
\label{tab:thresholds_filter}
\end{table}%

\begin{figure}[htp]
  \centering
  \includegraphics[width=0.98\linewidth]{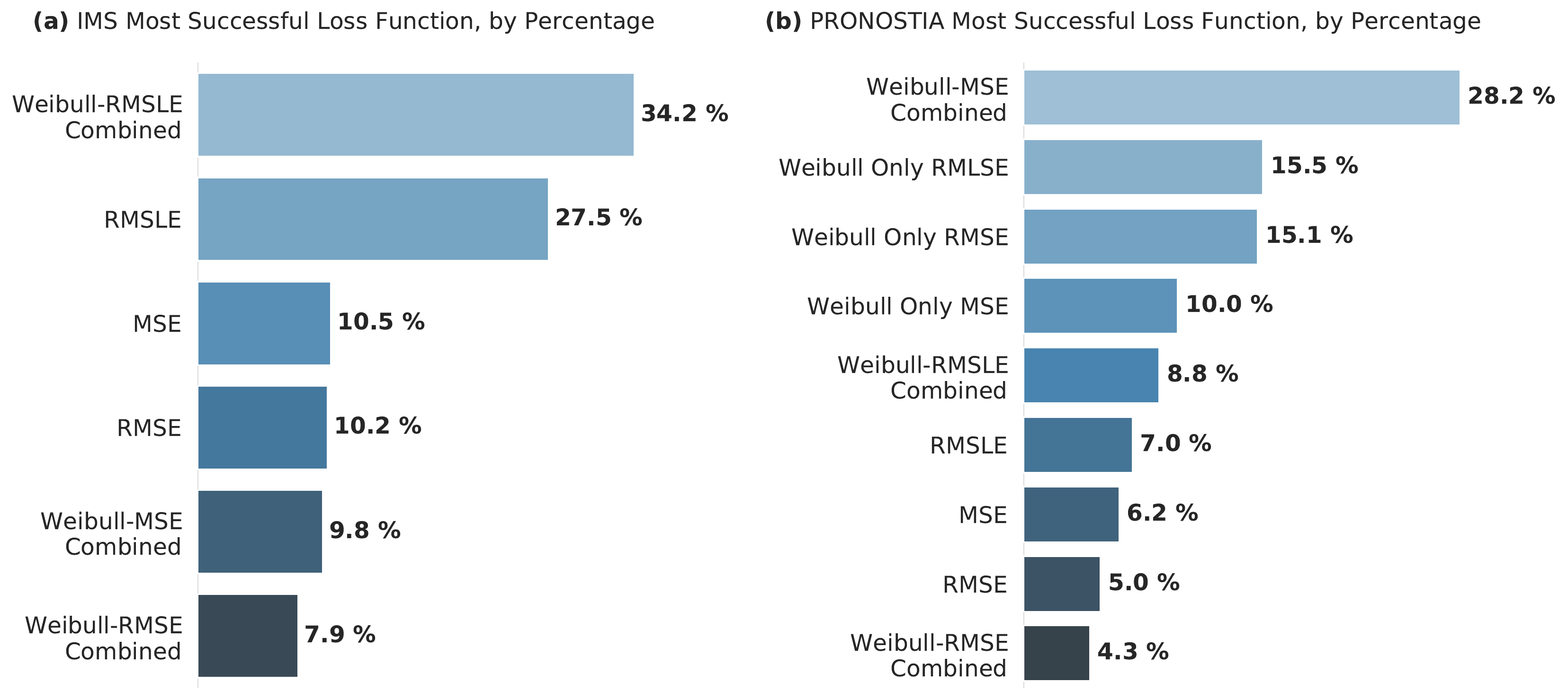}
  \caption{The loss functions, ranked by how frequently they appeared in the results, after the poorly performing models were filtered out.}
  \label{fig:percentages_loss_func}
 \end{figure}

After filtering, 775 unique model architectures remained from the IMS data set, and 582 from the PRONOSTIA data set. The top performing models, based on the $R^2$ from the test data, was selected from each of the 775 and 582 model architectures, respectively. Thus, it was found, empirically, which loss function was most effective, as shown in Figure~\ref{fig:percentages_loss_func}. The Weibull-RMSLE combined loss functions performed best on the IMS data set. Interestingly, nearly all the Weibull-based loss functions outperformed the traditional loss functions in the PRONOSTIA experiment.

\begin{figure}[htp]
  \centering
  \includegraphics[width=0.7\linewidth]{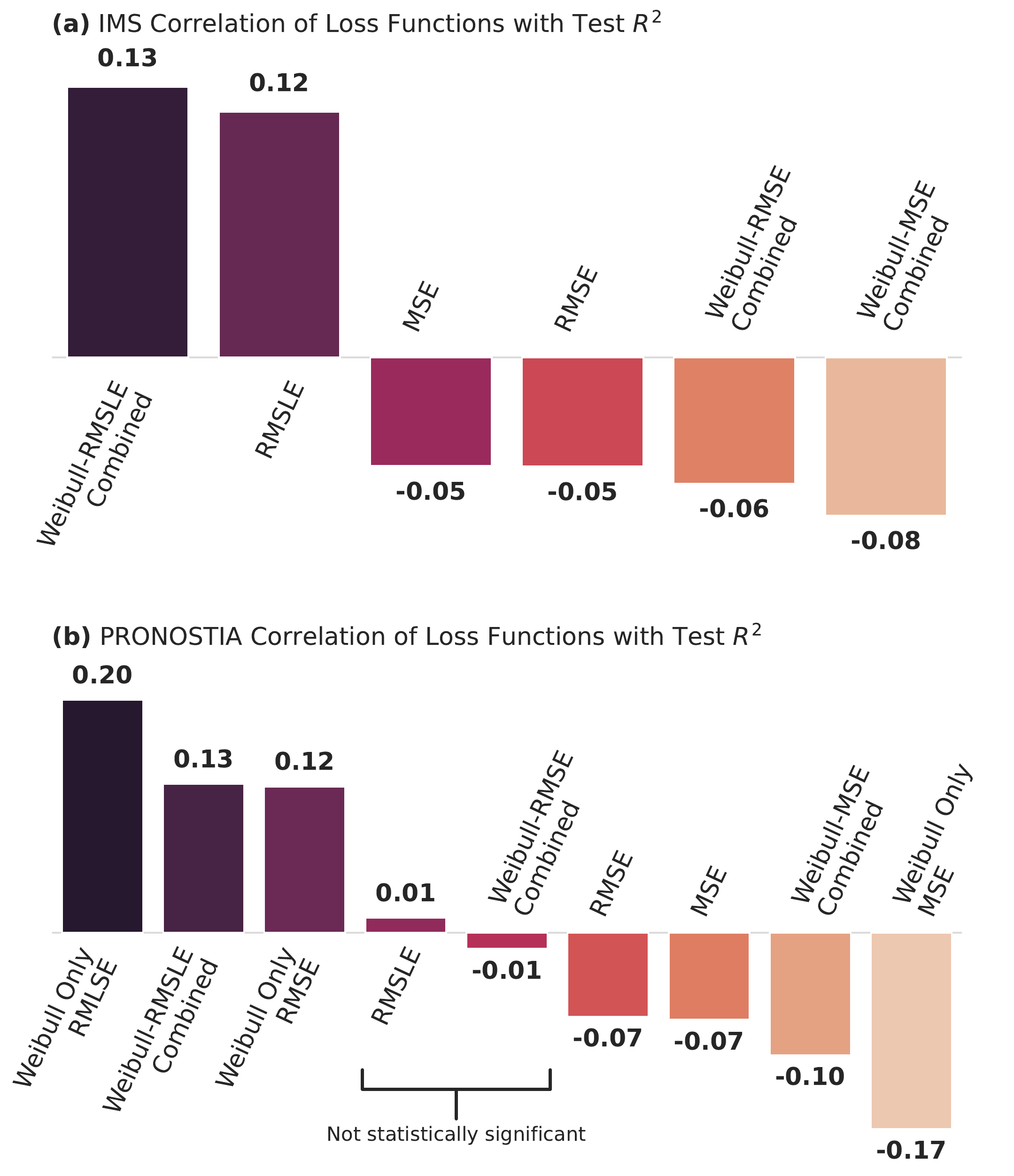}
  \caption{The loss functions, ranked by how frequently they appeared in the final results, after the poorly performing models were removed.}
  \label{fig:correlations}
 \end{figure}

To further evaluate the results, the point-biserial correlation coefficient of each loss function, against the test $R^2$, was found. Figure~\ref{fig:correlations} illustrates the correlation, after filtering, for both experiments. The results were found to be statistically significant (P-value < 0.05) for all the loss functions, except for the RMSLE and Weibull-RMSE combined on the PRONOSTIA data set. The ``Weibull only'' loss functions failed to produce any models that exceeded the filtering threshold in the IMS experiment, as shown in Figure~\ref{fig:correlations}-a. This result is further discussed in Section~\ref{further_analysis}.

\begin{figure}
  \centering
  \includegraphics[width=0.90\linewidth]{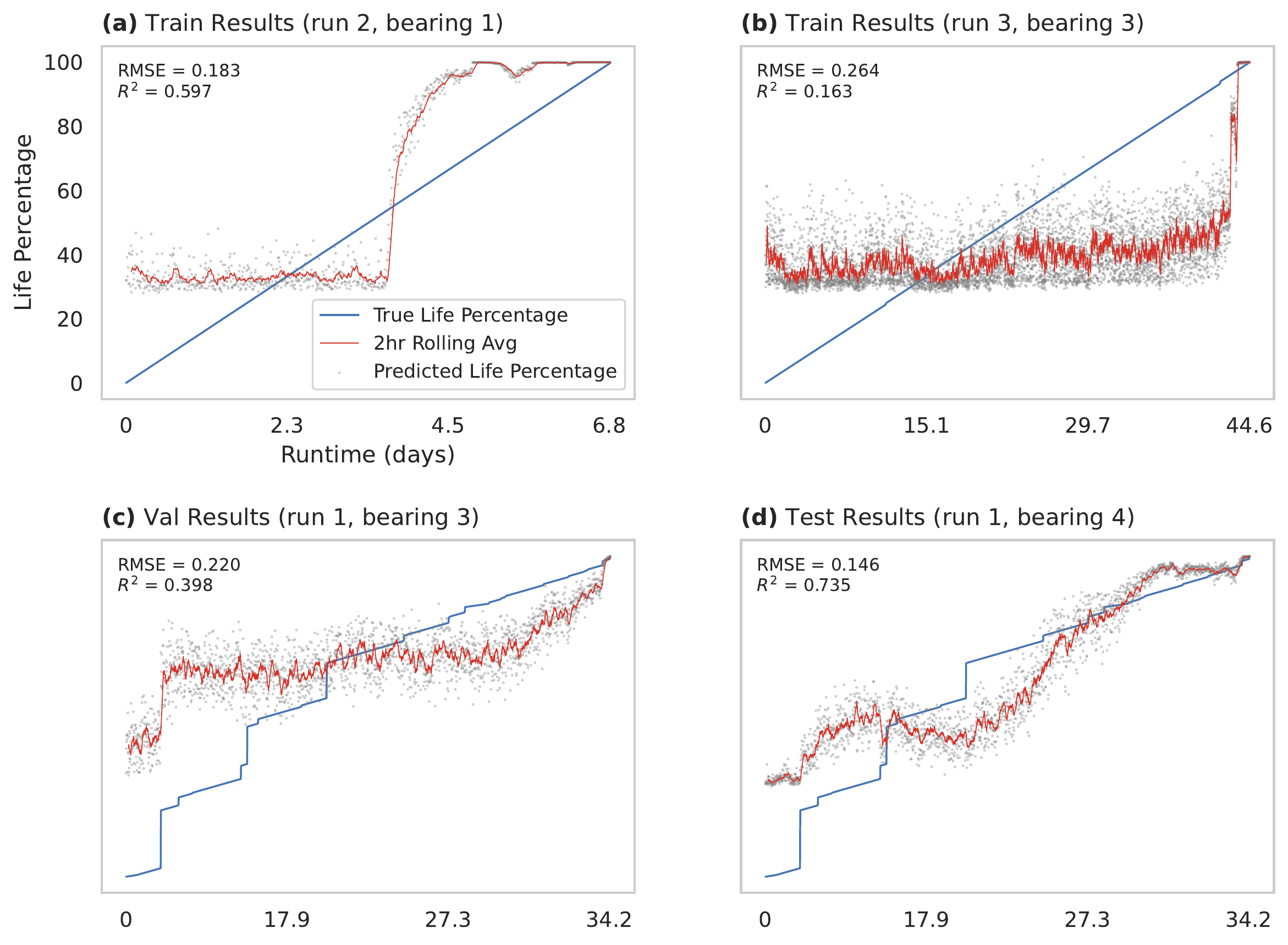}
  \caption{The predictions of the best performing model from the IMS experiment. The ``2hr rolling average'' is an average of the predictions over a 2-hour time frame and is added in order to clearly visualize the trend line. The ``jumps'' in the ``True Life Percentage'' line, such as in figure (c) are due to gaps in the collection of data. Note: the y-axis (life percentage) is the same for all four figures.}
  \label{fig:ims_results_plot}
 \end{figure}

\begin{figure}[htp]
  \centering
  \includegraphics[width=0.98\linewidth]{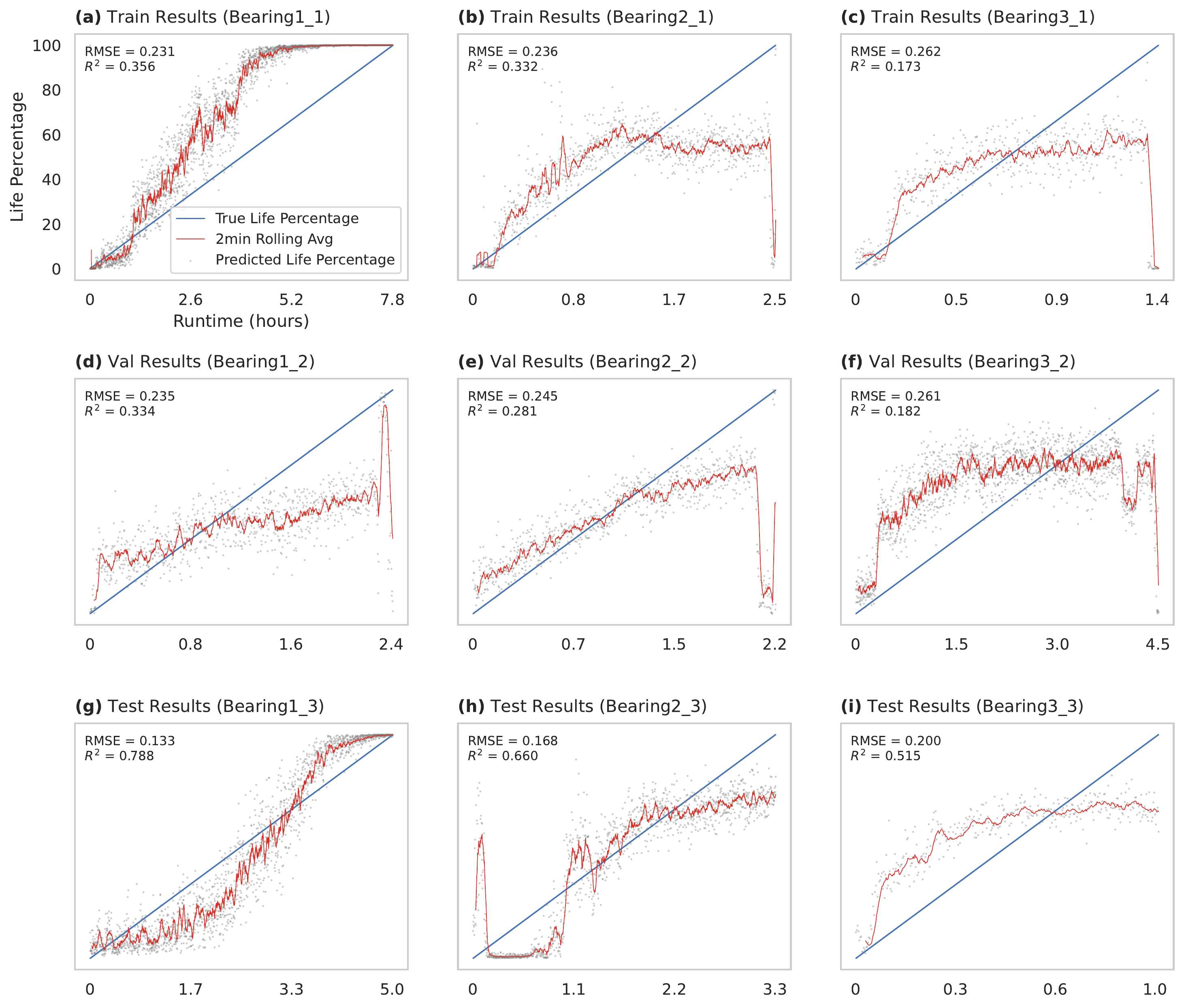}
  \caption{The predictions of the best performing model from the PRONOSTIA experiment. The ``2min rolling average'' is an average of the predictions over a 2-minute time frame and is added in order to clearly visualize the trend line.}
  \label{fig:pronostia_results_plot}
 \end{figure}

Table~\ref{tab:model_parameters} shows the parameters for the single best performing models from the IMS and PRONOSTIA experiments. Figures~\ref{fig:ims_results_plot} and~\ref{fig:pronostia_results_plot} show the prediction of these models across their respective training, validation, and testing data sets. Plotting the results, in this way, can assist the reader in gaining an intuitive understanding of the models’ performance. 

\begin{table}[htp]
\centering
\caption{The threshold values used to filter out poorly performing models.}
\begin{tabular}{lll}
\toprule
{} &                    Best IMS Model&            Best PRONOSTIA Model \\
\midrule
Loss Function   &  Weibull-RMSE Combined &  Weibull Only RMLSE \\
Layers          &                       4 &                    2 \\
Units per Layer &                      32 &                   32 \\
Drop Prob.      &                       0 &                 0.25 \\
$\lambda$               &                    0.53 &                 2.28 \\
$\beta$               &                       2 &                    2 \\
$\eta$               &               63.9 days &            4.8 hours \\
\bottomrule
\end{tabular}
\label{tab:model_parameters}
\end{table}%

Based on these results, the Weibull-based loss functions proves beneficial on the PRONOSTIA data set. In fact, only the Weibull-based loss functions (Weibull only RMSLE, Weibull-RMSLE combined, and Weibull only RMSE), as illustrated in Figure~\ref{fig:correlations}-b, convey a positive benefit to the predictive performance of the model. 

However, the results from the IMS experiment are less conclusive. The Weibull-RMSLE combined loss functions and the RMSLE loss functions, as shown in Figure~\ref{fig:correlations}-a, convey positive benefit to their models. Yet, the Weibull-RMSLE combined loss function is only slightly more effective than the RMSLE loss function. Furthermore, the Weibull only RMSLE loss function fails to produce any models that pass the filtering criteria. As such, one can conclude that the largest benefit comes from the use of the RMSLE metric, in both the Weibull-RMSLE combined and RMSLE loss functions, rather than the inclusion of the knowledge-based loss function. Further analysis and limitations are discussed below. 

\subsection{Limitations} \label{limitations}
As noted in the introduction, this research intentionally uses a simplified feature engineering and neural network approach and this may be seen as a limitation. However, the intent of the research is to demonstrate the effectiveness of the Weibull-based loss function as opposed to a state-of-the-art feature engineering or RUL technique. The simplified feature engineering and neural network approach allowed for many models to be trained, thus proving the effectiveness of the method with statistical significance. 

The above limitation also prevents the direct comparison of this method to other RUL techniques. However, the Weibull-based loss function would rarely be used in isolation. Rather, it would be used with a combination of techniques, such as advanced feature engineering, state-of-the-art deep learning networks and techniques, and data sets that are more robust. The Weibull-based loss function thus becomes one of many techniques that are combined, and adjusted, to make a viable RUL model. 

Another legitimate concern is the instability of the model, during training, when a shape parameter ($\beta$) other than 2.0 was used. The instability could indicate that an assumed $\beta$ of 1.5, in the case of the IMS data set, is unrealistic due to the contrived nature of the data. The instability could also indicate an architectural problem with the neural network. This problem may be rectified with standard tools of gradient clipping or changes to the learning rate. Further exploration is required, and other researchers are encouraged to modify the publicly available code to improve the methods. 

Finally, there are constraints in the data sets which lead to a difference in results between the PRONOSTIA and IMS experiments, which are discussed below. 

\subsection{Further Analysis} \label{further_analysis}
The Weibull-based loss functions is beneficial in the PRONOSTIA experiments. Yet, the Weibull-based loss function shows less benefit for the IMS experiments, and only when the RMSLE metric is used. Appreciating the challenges in the two data sets can provide possible explanations for the divergence in results.

Bearing manufacturers have worked over many years to produce reliable and long-life bearings. Given appropriate mechanical design and operating conditions (such as loading and lubrication), the life of a bearing should fall within a designated range, as noted by the bearings L10 life \cite{bloch2006maximizing}. Bearings designed and operated in these suitable conditions will naturally fail within a well parameterized Weibull distribution.

However, both the IMS and PRONOSTIA data sets are from test rigs that intentionally accelerate bearing failures through the application of high loads and stresses. These high loadings produce failure signatures that are abnormal from real-world applications.

In the PRONOSTIA data set, common failures signatures, such as ball pass frequencies, are not observable due to the excessive loading, as noted by the creators of the data set \cite{nectoux2012pronostia}. The excessive loading causes high levels of noise in the vibration signatures and a blending of many failure modes together. As such, it is reasonable to assume that the PRONOSTIA data set carries much less useful ``information'' in each of its runs than a comparable data set from a real-world application.

The lower information density in the PRONOSTIA data set makes the training of machine learning models more challenging. Thus, the machine learning models must rely on others sources of knowledge during training. In the case of our experiment, the knowledge derives from the introduction of the Weibull-based loss function. As such, the Weibull-based loss functions consistently outperform the traditional loss functions in our experiments, as shown in Figures~\ref{fig:percentages_loss_func}-b and \ref{fig:correlations}-b.

\begin{figure}[htp]
  \centering
  \includegraphics[width=1.0\linewidth]{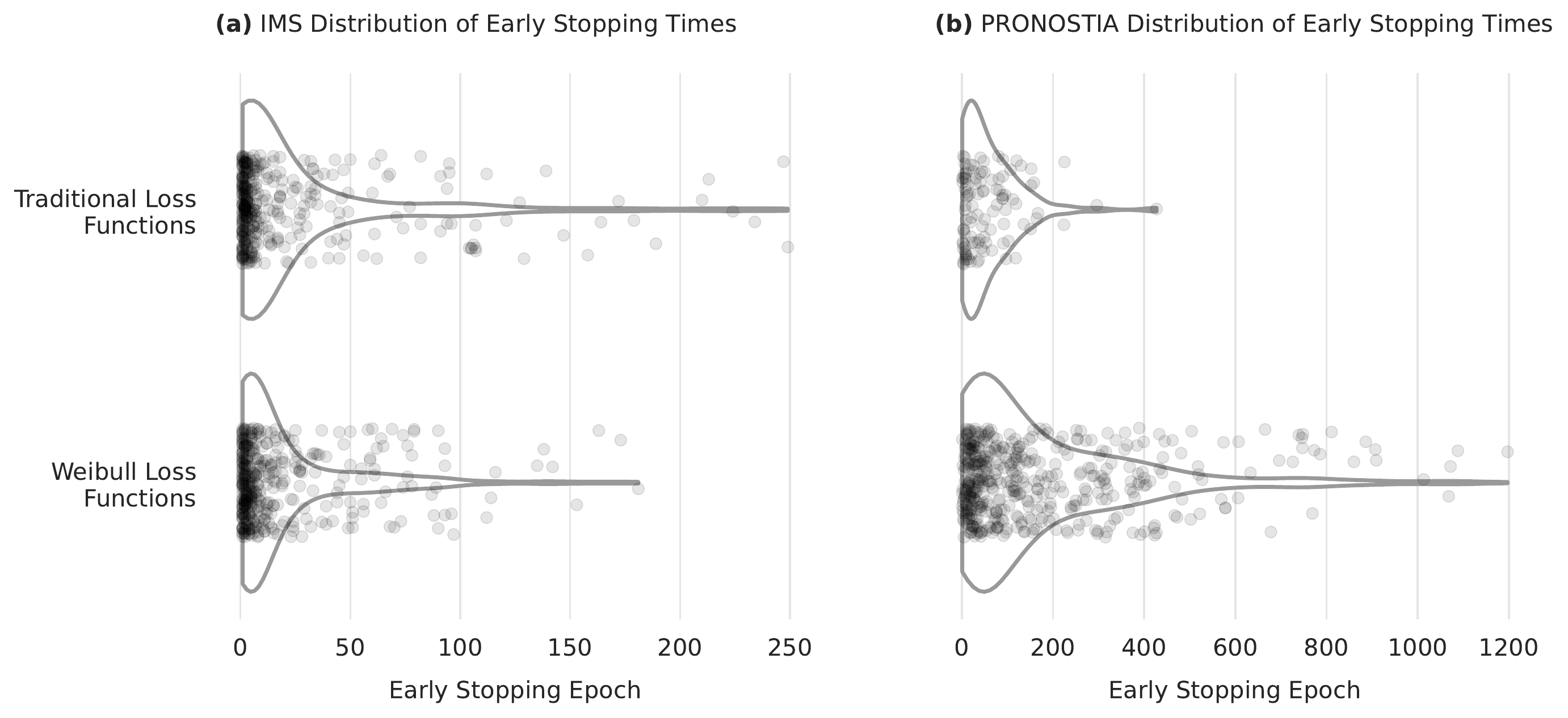}
  \caption{Distribution of early stopping times for both the IMS and PRONOSTIA experiments. Each dot represents when one model quit training due to early stopping. The width of the violin shape represents the relative frequency of early stopping times.}
  \label{fig:epoch_stop_dist}
 \end{figure}

The IMS data set better simulates a real-world industrial application and is of higher data quality. Ball pass frequencies are visible, demonstrating the superior informational quality in the data \cite{gousseau2016analysis}. However, the data set has fewer examples of failures than in the PRONOSTIA data set, and thus, less data for the machine learning models to learn from. Figure~\ref{fig:epoch_stop_dist} shows the distribution of early stopping times for both the IMS and PRONOSTIA experiments. The majority of models quit training after six epochs in the IMS experiments. For the PRONOSTIA data set, the majority of models stopped training between 36 and 98 epochs, depending on whether the Weibull-based loss function was used (see Table~\ref{tab:epoch_stop_stats} in the appendix for the summary statistics behind Figure~\ref{fig:epoch_stop_dist}). A neural network training on fewer examples will more quickly overfit the data, as is the case with the IMS experiment. Consequently, the early stopping engages sooner in the IMS experiments to prevent overfitting.

In addition, the limited number of runs in the IMS data set could produce a poor estimation of the characteristic life, $\eta$. The estimation error would also be affected by the inherently stochastic nature of bearing failures \cite{qiu2002damage}, which is not as present in the PRONOSTIA data set due to the blending of failure modes. Regarding the IMS data set, \citeauthor{gousseau2016analysis} \cite{gousseau2016analysis} speculate that the failure of bearing three, during run 3 (corresponding to Figure~\ref{fig:ims_results_plot}-b), is a ``false-positive''; that is, no failure takes place. Here, in opposition to \citeauthor{gousseau2016analysis}, we counter that the failure in this bearing likely exists. Rather, the failure is rapid in nature, again highlighting the stochasticity in bearing failures \cite{ISO2812007}. 

These factors, namely the lower informational quality of the PRONOSTIA data set, the limited data in the IMS data set, the potential error in the characteristic life, and the stochasticity in bearing failures, could explain the divergence in results between the IMS and PRONOSTIA experiments. 

The above discussion illustrates a pervasive problem in PHM research; that is, the limited availability of large and high-quality data sets \cite{wang2021recent}. Regardless, we believe the method presented here, through the Weibull-based loss function, will be useful to others. The process of integrating a Weibull-based loss function into a neural network is simple, especially when an accurate knowledge of the shape parameter already exists. The Weibull-based loss function, therefore, can be one of many parameters, along with regularization techniques and model hyperparameters, that can be tuned to produce optimal results. 

\subsection{Future Work}
A natural path for future work is to test the Weibull-based loss functions on data from real-world industrial environments. We suspect that applications consisting of large fleets of pumps or gas turbines, where failure modes and the associated Weibull distributions are well understood, would benefit from the use of the Weibull-based loss functions. 

More broadly, we are interested in exploring other methods of knowledge informed machine learning within PHM. The use of other constraints, such as the monotonicity and other approximation constraints, can be readily tested in the existing experimental framework presented here. The integration of physics-based models into the machine learner is also of interest. 

\section{Conclusion}
Integrating external knowledge into a machine learner, called knowledge informed machine learning, can improve the performance of PHM solutions. In this experiment, prior knowledge from the field of reliability engineering was used to improve the performance in a remaining useful life estimation task. The knowledge was represented through the probabilistic relationship of the Weibull distribution. The knowledge was integrated into the machine learning models using a Weibull-based loss function as an approximation constraint. 

The method was tested on the IMS and PRONOSTIA bearing data sets. In both experiments, the Weibull-based loss function provided benefits when compared to neural networks trained solely with traditional loss functions. However, the benefit was less significant on the IMS data set. Subjectively, the difference in results could derive from the lesser informational quality of the PRONOSTIA data set, and the relatively small size of the IMS data set.  

Integrating knowledge through the Weibull-based loss function shows promise. The Weibull-based loss function is simple to implement and can be combined with other techniques to improve neural network performance. All the code is publicly available to assist other researchers in their endeavors. 
 
\clearpage
\bibliographystyle{unsrtnat}
\bibliography{sources}

\small

\newpage
\section*{Appendix} \label{appendix}

\begin{figure}[h]
  \centering
  \includegraphics[width=1.0\linewidth]{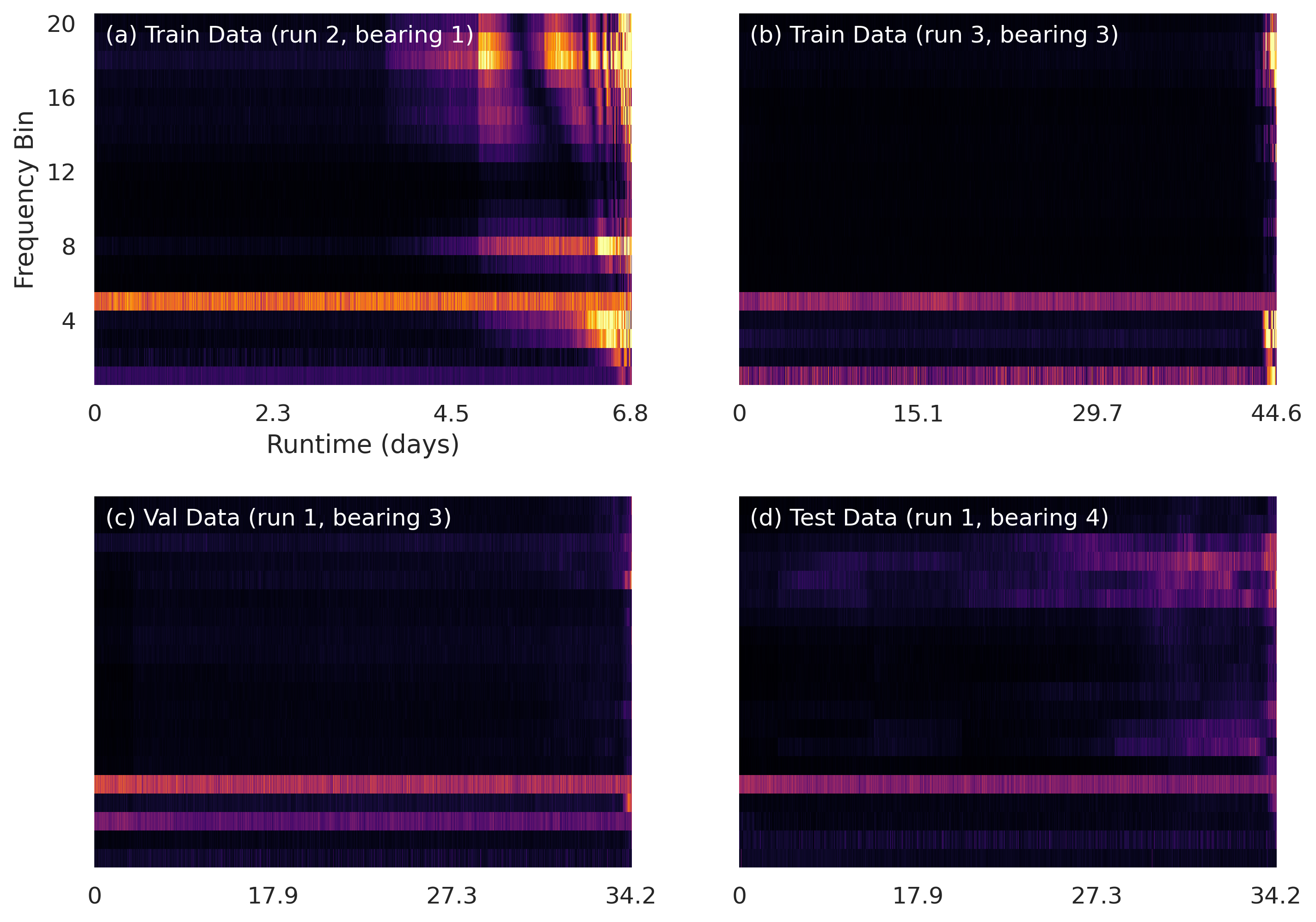}
  \caption{IMS spectrograms for the training, validation, and testing data after preprocessing. The color scale is the same for each spectrogram.}
  \label{fig:ims_spectrograms_appendix}
 \end{figure}
 
 \begin{figure}[h]
  \centering
  \includegraphics[width=1.0\linewidth]{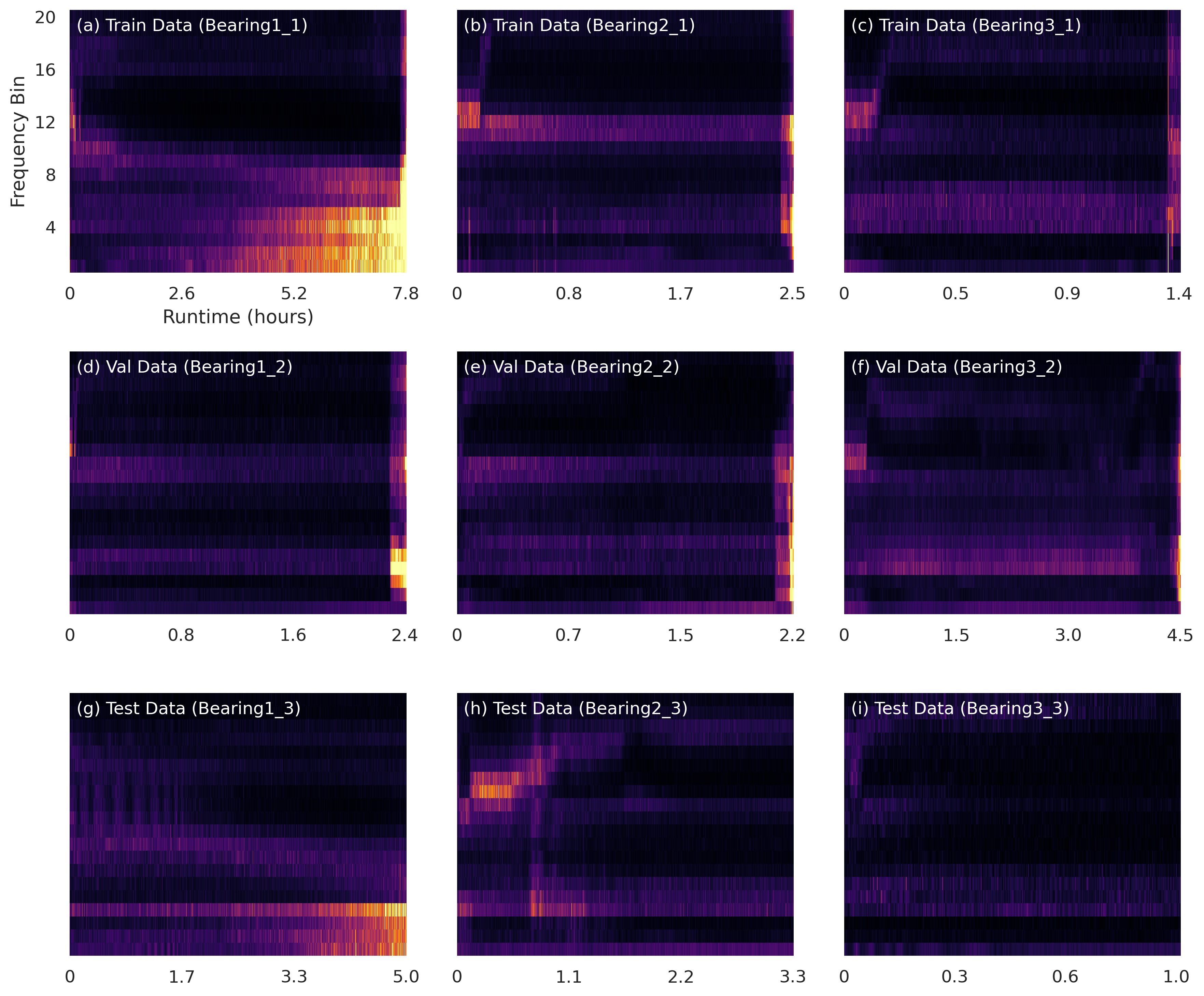}
  \caption{PRONOSTIA spectrograms for the training, validation, and testing data after preprocessing. The color scale is the same for each spectrogram.}
  \label{fig:femto_spectrograms_appendix}
 \end{figure}

 \begin{table}[hp]
\centering
\caption{Summary statistics for early stopping during the IMS PRONOSTIA experiments. Statistics are taken after filtering of poorly performing models.}
\begin{tabular}{crr|rr} 
\toprule
      & \multicolumn{2}{c}{IMS Experiment} & \multicolumn{2}{c}{PRONOSTIA Experiment}  \\ 
\midrule
      & Traditional Loss & Weibull Loss    & Traditional Loss & Weibull Loss           \\ 
\midrule
count & 373~             & 402~            & 106~             & 476~                   \\
mean  & 32.8~            & 23.4~           & 57.2~            & 219.7~                 \\
std   & 79.6~            & 46.8~           & 66.6~            & 333.6~                 \\
min   & 1~               & 1~              & 1~               & 1~                     \\
25\%  & 2~               & 2~              & 9.25~            & 30~                    \\
50\%  & 5~               & 6~              & 36.5~            & 98~                    \\
75\%  & 25~              & 22.75~          & 89~              & 273~                   \\
max   & 665~             & 367~            & 427~             & 1950~                  \\
\hline
\end{tabular}
\label{tab:epoch_stop_stats}
\end{table}

\end{document}